\documentclass[lettersize,journal]{IEEEtran}
\usepackage{stfloats}
\usepackage{url}
\usepackage{verbatim}
\usepackage{cite}
\usepackage{amsfonts}
\usepackage{algorithmic}
\usepackage{graphicx}
\usepackage{lipsum}
\usepackage{mathtools, lipsum, cuted}
\usepackage{mathptmx}          
\usepackage{helvet}            
\usepackage{courier}           
\usepackage{type1cm}           
\usepackage{makeidx}           
\usepackage{graphicx}          
\usepackage{float}             
\usepackage{url}
\usepackage{gensymb}
\usepackage{multicol}   
\usepackage{lipsum}   
\usepackage[bottom]{footmisc}  
\usepackage{array,multirow}
\usepackage{amsmath,amssymb,dsfont}
\usepackage{amsfonts}
\usepackage{textcomp}
\usepackage[caption=false]{subfig}
\usepackage{bm}
\usepackage{eucal}
\usepackage{algorithm2e}
\usepackage{latexsym}
\usepackage{epsfig}
\usepackage{amstext}
\usepackage{caption}
\usepackage{balance}
\DeclareMathOperator*{\argmax}{argmax} 

\def\BibTeX{{\rm B\kern-.05em{\sc i\kern-.025em b}\kern-.08em
    T\kern-.1667em\lower.7ex\hbox{E}\kern-.125emX}}
\begin{document}
\title{On Scale Space Radon Transform, Properties and Application in CT Image Reconstruction}
\author{N. Nacereddine, D. Ziou, and A.B. Goumeidane
\thanks{Manuscript created Month, ??th 2023}
\thanks{N. Nacereddine and A.B. Goumeidane are with Research Center in Industrial Technologies CRTI, P.O.Box 64, Ch\'{e}raga, Algiers 16014, Algeria (e-mails: n.nacereddine@crti.dz ; a.goumeidane@crti.dz).}
\thanks{D. Ziou is with DMI, Universit\'{e} de Sherbrooke, Qu\'{e}bec, QC J1K 2R1, Canada (e-mail: djemel.ziou@usherbrooke.ca).}}

\markboth{IEEE Transactions on \dots,~Vol.~??, No.~??, Month~2023}%
{How to Use the IEEEtran \LaTeX \ Templates}

\maketitle

\begin{abstract}
Since the Radon transform (RT) consists in a line integral function, some modeling assumptions are made on Computed Tomography (CT) system, making image reconstruction analytical methods, such as Filtered Backprojection (FBP), sensitive to artifacts and noise. In the other hand, recently, a new integral transform, called Scale Space Radon Transform (SSRT), is introduced where, RT is a particular case. Thanks to its interesting properties, such as good scale space behavior, the SSRT has known number of new applications. In this paper, with the aim to improve the reconstructed image quality for these methods, we propose to model the X-ray beam with the Scale Space Radon Transform (SSRT) where, the assumptions done on the physical dimensions of the CT system elements reflect better the reality. After depicting the basic properties and the inversion of SSRT, the FBP algorithm is used to reconstruct the image from the SSRT sinogram where the RT spectrum used in FBP is replaced by SSRT and the Gaussian kernel, expressed in their frequency domain. PSNR and SSIM, as quality measures, are used to compare RT and SSRT-based image reconstruction on Shepp-Logan head and anthropomorphic abdominal phantoms. The first findings show that the SSRT-based method outperforms the methods based on RT, especially, when the number of projections is reduced, making it more appropriate for applications requiring low-dose radiation, such as medical X-ray CT. While SSRT-FBP and RT-FBP have utmost the same runtime, the experiments show that SSRT-FBP is more robust to Poisson-Gaussian noise corrupting CT data.
\end{abstract}



\section{Introduction}
The Radon transform (RT) \cite{Deans83} is the main mathematical tool used in Computed Tomography (CT) imaging where, the transform invertibility has made it extremely useful, involving image reconstruction from projections. Major image reconstruction algorithms fall into two categories: (1) the analytical methods based directly on the RT inverse formula, such as filtered backprojection (FBP) \cite{Dud84,Kak88} and (2) the algebraic reconstructions techniques which are, generally, iterative, alternating between the projection and image domains, involving the re-projection of an estimated image \cite{Gal97}.
Even if analytical methods are computationally efficient and easy to implement, they require excessive projections to reliably reconstruct high-quality images. In addition, analytical methods generally ignore measurement noise in the problem formulation and treat noise-related problems as an "afterthought" by post-filtering operations \cite{Fessler09}. This is considered as limitations of analytical reconstruction methods that impair their performance and restrict their utilization in radiation sensitive applications, including biological tissue studies \cite{Wang18}.

To address the drawback of the analytical methods, algebraic methods based on iterative algorithms have been extensively studied. Compared to analytical reconstruction methods, iterative techniques have the major advantage that they permit the emission and detection process to be accurately modeled and prior knowledge is relatively easily incorporated in the inverse process using penalty functions. They can also be performed with projections of limited number, or not uniformly distributed over 180 or 360 degrees \cite{Kak88}. A first method, called algebraic reconstruction technique (ART), which was introduced in 1970 by Gordon et al. \cite{Gordon70}, had a relatively rapid convergence speed but suffered from heavy salt and pepper noise. Other improved method versions, such as simultaneous iterative reconstruction technique (SIRT) \cite{Gilbert72,Gregor08}, simultaneous algebraic reconstruction technique (SART) \cite{Andersen84,Jiang03} and recent iterative approaches applying regularization term, are proposed to reduce the noise but are, all, time consuming. To sum up over the algebraic methods, these are more robust to incomplete and noisy projections, but they have the disadvantage of high computational cost.

When both the analytic and iterative methods are numerically implemented, both the forward and back projection operations play a primary role in the overall computational process \cite{Yu12}.

There are many methods to model the forward and back projection techniques for a discrete imaging object. All of those models compromise between computational complexity and accuracy \cite{Miao14}. Among these models, one can cite pixel-driven model (PDM) \cite{Peters81}, ray-driven model (RDM) \cite{Zeng93}, distance-driven model (DDM) \cite{DeMan02}, area-integral model (AIM) \cite{Yu12} and improved distance-driven model (IDDM) \cite{Miao14}. 

As mentioned at the beginning of this section, the Radon transform, which is a line integral, is the main tool used in CT imaging. However, in order to make the mathematics manageable, during the derivation of FBP algorithm, a series of assumptions are made: (1) the physical dimension of the X-ray focal spot is assumed to be infinitely small, approximating a point source, (2) all X-ray photon interactions are assumed to take place at the geometric central point of the detector cell and (3) the shape ad size of image voxel are neglected and only the geometric center of the voxel is considered \cite{Hsieh13}.  

However, these assumptions do not describe the reality and, for most clinical CT scanners, finite dimensions of the evoked CT system elements are considered where, for example, typical focal spot size is about 1 mm $\times$ 1 mm and detector cell spacing is about 1.4 mm with an active area of 1.1$\sim$ 1.2 mm. The image voxel is related to both the pixel size and slice thickness where the former is dependent on both the field of view and the image matrix. The matrix size is typically $256\times 256$ or $512\times 512$ while, the pixel size is typically between 0.5 and 1.5 mm.

Introduced recently in \cite{Ziou21}, the Scale Space Radon Transform consists in a matching of an embedded object in an image and the Gaussian Kernel parametrized by a scale parameter $\sigma$ where, the abovementioned CT system component dimensions, including the focal spot size, the voxel dimensions and the detector width, could be adequately approximated during forward and back projection operations. 

In this work, as an alternative to the abovementioned iterative construction techniques which are time consuming, a new version of Filtered Backprojection based on SSRT, named SSRT-FBP, is proposed. In addition to its fast implementation, the SSRT-FBP reconstruction technique is robust to noise and reduced number of projections thanks to the nice behavior of SSRT in scale space.
      
SSRT is a full-fledged integral transform where, RT is a particular case, obtained when the Gaussian kernel is replaced by a Dirac distribution $\delta$. Although its recency, the SSRT can be used for a wide range of computer vision applications \cite{Ziou21,Gou22}. Like any other integral transform, SSRT has several properties, some of which will be used by the FBP algorithm. In this paper, some basic properties of SSRT will be presented. Furthermore, the approaches of SSRT inversion will be developed and discussed where, the new version of FBP will be derived in detail and applied on reconstruction of test and real CT scan images. 

The remainder of the paper is organized as follows. The preliminaries and work motivation are depicted in Section 2. Section 3 is devoted to the SSRT properties including its inversion. Image reconstruction using SSRT is detailed in Section 4 and main experiments and results are reported in Section 5. Lastly, main conclusions and perspectives are drawn in Section 6. 

\section{Preliminaries and motivation}
\subsection{SSRT definition}
\begin{figure}[!tbh]
\centering
\includegraphics[scale=0.5]{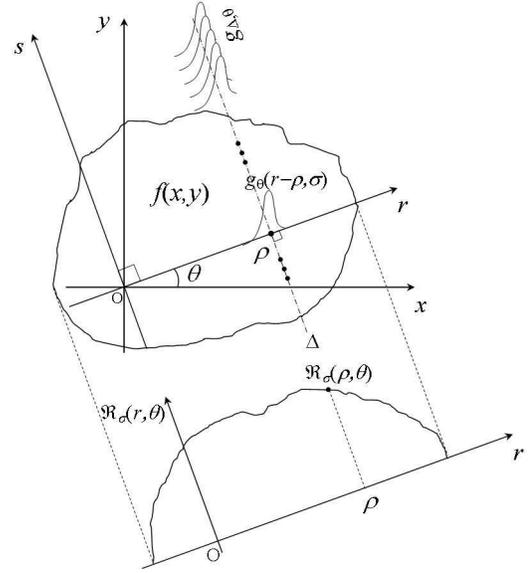}
\caption{Definition of the Scale Space Radon Transform} 
\label{ssrt_def}
\end{figure}

Let $(x,y)$ the Cartesian coordinates of a point in a 2-D space, $f(x,y)$ a continuous function with a compact support and $g_{\Delta,\theta}$ a set of parallel linear Gaussians $g_\theta(r-\rho,\sigma)$, parameterized by the orientation angle $\theta$, the standard deviation $\sigma$ and the centerline $(\Delta)$: $x\cos\theta+y\sin\theta-\rho=0$, as illustrated in Fig.~\ref{ssrt_def}. By referring to Fig.~\ref{ssrt_def}, Scale Space Radon transform (SSRT), noted $\Re_{\sigma}f(\rho,\theta)$, is defined by \cite{Ziou21}

\begin{equation}
\Re_{\sigma}f(\rho,\theta) =\int_{-\infty}^{+\infty}\int_{-\infty}^{+\infty}f(x,y)g(x\cos\theta + y \sin\theta-\rho) dx dy
\end{equation}
\label{ssrt_def_eq}
where, $g(t)=\frac{1}{\sqrt{2\pi}\sigma}e^{\frac{-t^2}{2\sigma^2}}$ is the Gaussian function.

To establish the relationship between SSRT and RT, let consider a $(r,s)$ coordinate system, as in Fig.~\ref{ssrt_def}, a rotated version of the original $(x,y)$ system, expressed as 
\begin{equation}
\left[\begin{array}{c} r \\ s \end{array}\right] =
\left[\begin{array}{cc} \cos\theta & \sin\theta\\ -\sin\theta & \cos\theta\end{array}\right]
\left[\begin{array}{c} x \\ y \end{array}\right] 
\end{equation}
It follows that the SSRT is rewritten as
\begin{equation}
\Re_{\sigma}f(\rho,\theta) =\int_{r}\int_{s}f(r\cos\theta-s\sin\theta,r\sin\theta+s\cos\theta)g(r-\rho) dr ds
\end{equation}
Equivalently,

\begin{equation}
\Re_{\sigma}f(\rho,\theta) = \int_{r}g(r-\rho)\left[\int_{s}f(r\cos\theta-s\sin\theta,r\sin\theta+s\cos\theta)ds\right]dr
\end{equation}

Note that the expression between brackets is nothing but the expression of the Radon transform $\mathcal{R}f(\rho,\theta)$.
It follows that
\begin{equation}
\label{ssrt_conv}
\Re_{\sigma}f(\rho,\theta) = \int_{r}g(r-\rho)\mathcal{R}f(r,\theta)dr = (g\ast^\rho\mathcal{R}f)(\rho,\theta)
\end{equation}
where $\ast^\rho$ is the convolution operation with regards to (w.r.t.) the variable $\rho$.

The formulae (\ref{ssrt_conv}) constitutes a straightforward way to compute the SSRT which is nothing but the convolution of Radon transform with a Gaussian kernel tuned by a scale parameter $\sigma$. It is important to note that the formulae (\ref{ssrt_conv}) could not be derived from the Radon transform Convolution Theorem (RCT) since, here, it consists in a convolution of an Gaussian function represented in the spatial plane $(r,s)$ and not in Radon space $(\rho, \theta)$, as stated in RCT. Consequently, all the properties, given in the next section, are specific to SSRT and could not be obtained, under any circumstances, from the RCT.

\subsection{Motivation of SSRT utilization}
In Computed Tomography, three problems have to be addressed: how to model the discrete representation of the continuous object to scan, how to model the forward and back projections operation and how to practically implement it efficiently and accurately \cite{Lev14}.
Before to discuss on the question of modeling used in CT reconstruction, let give the principle of the latter. The CT reconstruction problem may be stated as follows: given a finite set of projections $p_i$ ($i=1,2,\dots,M$ for $M$ individual measurements) of a 2D (or 3D) function $f(x,y)$ (or $f(x,y,z)$) with compact support, representing the X-ray attenuation coefficients, obtain the best estimate of that function. Each projection may generally be written as a weighted 2D (or 3D) integral of the function $f$

\begin{equation}
\label{gen_proj}
p_i=\int_{-\infty}^{+\infty}\int_{-\infty}^{+\infty}K_i(x,y)f(x,y)dxdy
\end{equation}
where, $K_i(x,y)$ are the weighting functions related to the projection geometry. 

The discrete approximation of $f(x,y)$, noted $\bar{f}(x,y)$, consists of a set of samples of the continuous image $f(x,y)$ over a uniform grid which can be represented as the superposition of scaled and shifted copies of the basis functions $\varphi$, 

\begin{equation}
\label{discr_func}
\bar{f}(x,y)=\sum_{j=1}^{N}c_j \varphi(x-x_j,y-y_j)
\end{equation}

where, $\varphi(x,y)$ is the basis function centered at the origin, $\{c_j\}_{j=1}^N$ is the set of coefficients of the image representation and $\{(x_j,y_j)\}_{j=1}^N$ be a set of $N$ points in 2D space that are the nodes of the uniform grid with node spacing $\Delta$, assumed equal, in general, to the unity.

The common model for the basis functions for 2D (or 3D) space is the pixel (or voxel) basis function which is a superposition of non-overlapping pixels (or voxels) of uniform size, i.e. in the case of a pixel, a basis with value 1 for $|x|,\ |y|\leq\Delta/2$ and value 0 otherwise \cite{Lev14,Han85}.
Similarly to voxels, another basis function for image representation using a spatially localized and spherically symmetric volume elements (blob) based on Kaiser-Bessel window functions, is considered \cite{Lew92,Matej96}. 
An advantage of spherically symmetric basis functions is their identical appearance, independent of the projection angle, allowing efficient calculation of line integrals. A smooth bell-shaped radial profile and the overlapping nature of blobs allows for creating a smooth representation of naturally smooth biological objects. However, the overlapping degree increases the computational complexity \cite{Lev14}.
That is said, the choice of the representation model of the object to scan is important because it influences the accuracy of the reconstructed image and the speed and the computational complexity of the algorithm.
  
Once the continuous object is discretized, its expression in (\ref{discr_func}) is substituted in (\ref{gen_proj}), which leads to  

\begin{equation}
\label{gen_proj_bis}
p_i=\sum_{j=1}^{N}c_j\int_{-\infty}^{+\infty}\int_{-\infty}^{+\infty}K_i(x,y)\varphi(x-x_j,y-y_j)dxdy
\end{equation}

Typically, the $K_i$ have large values within a narrow strip and small or zero values outside the strip. If the $K_i$ are unity within a straight strip of constant width and zero outside, (\ref{gen_proj_bis}) becomes a strip integral. 

For zero strip width, it becomes a line integral, called Radon transform where, the straight line $L_{\theta}$ is parameterized by the signed distance $p$ to an origin and an angle $\theta$ with the $x$-axis. In which case $K_i(x,y)= x\cos\theta_i+y\sin\theta_i-\rho_i$ and the line integral can be expressed as

\begin{equation}
\label{dig_rt}
p_i^{(L)}=\sum_{j=1}^{N}c_j\int_{L_{\theta_i}}\varphi(x-x_j,y-y_j)dxdy
\end{equation}

In many applications, the measurements are assumed to be approximated by line integrals. However, all real measurement systems including CT system measure along strips with a finite width. There are two reasons to optimize these beam widths: signal level and minimization of spatial aliasing \cite{Ing08}. As evoked in the Introduction, the finite dimensions of the CT system components such as X-ray source focal spot, object voxel and detector cell require that numerical approximations to be made on the projection and back projection calculation models during image reconstruction process.  

While remaining in the spirit of the strip integral mentioned above, our work consists in using a strip whose values are not equal to unity but follow a Gaussian profile for each slice perpendicular to the strip centerline. Then, in this case, the integration will be done on the object along a Gaussian-profile strip denoted $g_{\sigma,\theta}$, as shown in Fig.~\ref{ssrt_def}. 
For a given standard deviation $\sigma$, with slope $\theta$, we define Gaussian-profile strip integral, named SSRT, as follows:

\begin{equation}
\label{dig_SSRT}
p_i=\sum_{j=1}^{N}c_j\int_{(x,y)\in{g_{\sigma,\theta_i}}}\varphi(x-x_j,y-y_j)dxdy
\end{equation}

Let us examine the SSRT for a unit pixel in order to see the influence of SSRT on the basis function $\varphi$ defined as $\varphi(x,y)=1 \ if \ x,y\in[-0.5,0.5]$ and $\varphi(x,y)=0 \ elsewhere$. In these conditions, if we take $N=1$ and $c_j=1$ and if we assume that the Gaussian-profile strip centerline passes through the space origin $(0,0)$ giving $\rho_i=0$, then, we obtain

\begin{equation}
\label{ssrt_basis_x_y}
p_{i,\varphi}=\int_{-\infty}^{+\infty}\int_{-\infty}^{+\infty}g_\sigma(x\cos\theta_i+y\sin\theta_i)\varphi(x,y)dxdy
\end{equation} 

which, by using (\ref{ssrt_conv}), can be expressed as

\begin{equation}
\label{ssrt_basis_r_thet}
p_{i,\varphi}=\int_{r}^{}g_{\sigma}(r)\mathcal{R}_\varphi(r,\theta_i)dr
\end{equation}  

where, $\mathcal{R}_\varphi$ is the line integral of the unit pixel, in function of the radius $r$, and $g_{\sigma}(r)$ is the Gaussian function with zero mean and $std$ $\sigma$. 
The function $\mathcal{R}_\varphi$ is nothing but the RT of unit square, centered at the $x-y$ coordinate system origin $(0,0)$, of which the values, according to $\theta$, are given as follows \cite{Nac14}:

\begin{flushleft}
\small 
for $0\leq\theta\leq\frac{\pi}{4}$
\[
\mathcal{R}_\varphi(r,\theta) = \left\{{\begin{array}{ll}
1/\cos\theta & \mathrm{ if \ } 0\leq r \leq \frac{\cos\theta-\sin\theta}{2}\\
\frac{\cos\theta+\sin\theta-2r}{\sin 2\theta} & \mathrm{ if \ } \frac{\cos\theta-\sin\theta}{2}\leq r\leq
\frac{\cos\theta+\sin\theta}{2}
\end{array}}\right.
\]
for $\frac{\pi}{4}\leq\theta\leq\frac{\pi}{2}$
\[
\mathcal{R}_\varphi(r,\theta) = \left\{{\begin{array}{ll}
1/\sin\theta & \mathrm{ if \ } 0\leq r \leq \frac{\sin\theta-\cos\theta}{2}\\
\frac{\cos\theta+\sin\theta-2r}{\sin 2\theta} & \mathrm{ if \ } \frac{\sin\theta-\cos\theta}{2}\leq r\leq
\frac{\cos\theta+\sin\theta}{2}
\end{array}}\right.
\]
for $\frac{\pi}{2}\leq\theta\leq\frac{3\pi}{4}$
\[
\mathcal{R}_\varphi(r,\theta) = \left\{{\begin{array}{ll}
1/\sin\theta & \mathrm{ if \ } 0\leq r \leq \frac{\cos\theta+\sin\theta}{2}\\
\frac{\cos\theta-\sin\theta+2r}{\sin 2\theta} & \mathrm{ if \ } \frac{\cos\theta+\sin\theta}{2}\leq r\leq
\frac{\sin\theta-\cos\theta}{2}
\end{array}}\right.
\]
for $\frac{3\pi}{4}\leq\theta\leq\pi$
\[
\mathcal{R}_\varphi(r,\theta) = \left\{{\begin{array}{ll}
-1/\cos\theta & \mathrm{ if \ } 0\leq r \leq -\frac{\cos\theta+\sin\theta}{2}\\
\frac{\cos\theta+\sin\theta-2r}{\sin 2\theta} & \mathrm{ if \ } -\frac{\cos\theta+\sin\theta}{2}\leq r\leq
\frac{\sin\theta-\cos\theta}{2}
\end{array}}\right.
\]
\end{flushleft}
\normalsize

\begin{figure}
\centering
\includegraphics[scale=0.35]{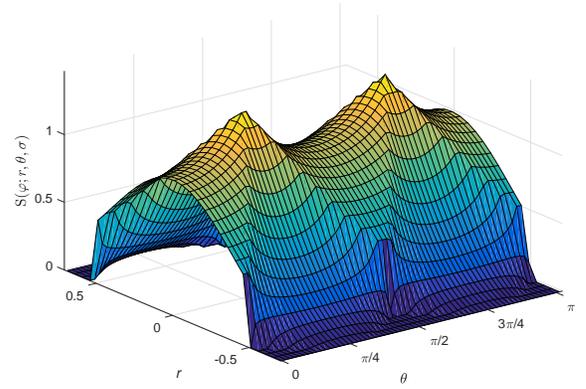}
\caption{Amplitude of SSRT-based pixel footprint for $\sigma=0.4$} 
\label{ssrt_ftprt_fig}
\end{figure}

\begin{figure}
\centering
\includegraphics[scale=0.5]{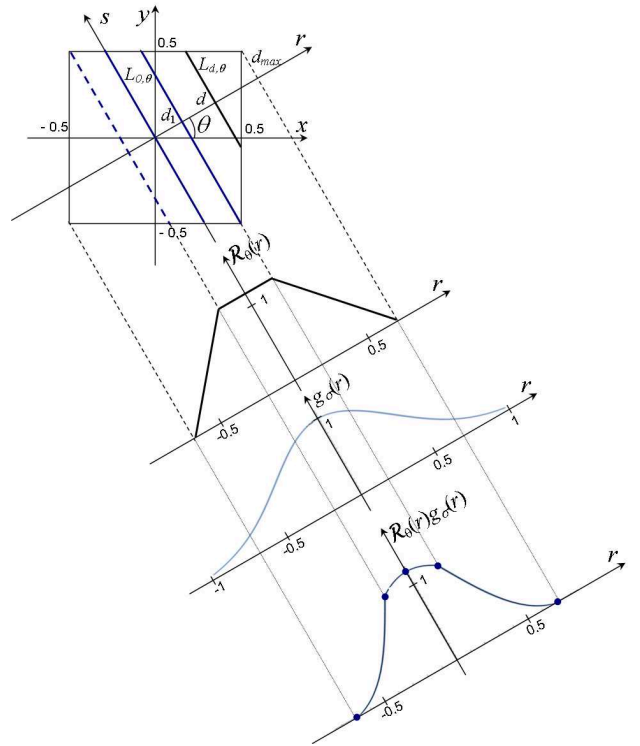}
\caption{SSRT-based pixel footprint for $\sigma=0.4$ and $\theta= \pi/6$} 
\label{ssrt_ftprt_theta}
\end{figure}

So, by using a compact formulation of the unit square RT $\mathcal{R}_\varphi(r,\theta)$, derived from its case-wise results, the analytical expression of the SSRT-based pixel footprint $S(\varphi;r,\theta,\sigma)=g_{\sigma}(r)\mathcal{R}_\varphi(r,\theta_i)$ (a function of the radius $r$), for each projection angle $\theta (\theta\in[0\ \pi])$, is obtained as follows:

\small
\begin{align}\nonumber
\label{ssrt_ftprt}
S(\varphi;r,\theta,\sigma)& = \frac{\exp\left[-\frac{r^2}{2\sigma^2}\right]}{2\sqrt{2\pi}\sigma\sin 2\theta}\bigg(\big|\sin\theta+\cos\theta-2r\big|+\big|\sin\theta+\cos\theta+2r\big|\\ &-\big|\sin\theta-\cos\theta-2r\big|-\big|\sin\theta-\cos\theta+2r\big|\bigg)
\end{align}
\normalsize
 
where, the parameter $\sigma$ controls the shape of the SSRT-based basis function. The higher the value of this parameter, the wider the shape of the SSRT-based pixel footprint and vice versa, where, especially, a value of $\sigma$ close to zero leads to point projection of the pixel. A 3D representation of the SSRT-based pixel projection given in (\ref{ssrt_ftprt}), for $\sigma=0.4$ and for all $\theta$, is illustrated in Fig.~\ref{ssrt_ftprt_fig}; whereas, an example, for $\theta=\pi/6$, is depicted in Fig.~\ref{ssrt_ftprt_theta}. This figure is an illustration of the SSRT at the level of a pixel basis function, which gives a $\theta$-oriented projection of the unit pixel weighted by a Gaussian function where, the largest weight is assigned at the pixel center, corresponding to the detector cell center. At the level of the entire image, representing the scanned object, the Gaussian weighting function is expressed as $w_{i,j}=g_\sigma(x_j\cos\theta_i+y_j\sin\theta_i-\rho_i)$ for the pixel $(x_j,y_j)$ and the projection value $p_i$ of $i$th detector. 

Then, the formula (\ref{dig_SSRT}) can be rewritten as $p_i=\sum_{j=1}^N c_j w_{i,j}$, consisting in the SSRT-based forward projection model, which does not require any further interpolation or approximation operations, unlike pixel-driven, ray-driven, distance-driven or area-integral models, cited in the Introduction. These coarse approximations made by such models are time cnsuming anf give rise to artifacts in the reconstructed object unless a great deal of projections are used.

Smoother and more isotropic basis functions better represent continuous functions and provide simpler projectors. It is the case of the blobs, already, explained at the beginning of the current subsection. Set apart their isotropy, blobs are often considered too computationally expensive in practice since the overlapping nature of blobs has to be taken into account. 

For the SSRT-based basis function, it is worth to note that the effect of the anisotropic behavior of the cubic voxel (or the square pixel) has been somewhat attenuated with the Gaussian function, as shown in Fig.~\ref{ssrt_ftprt_fig} where, the function $S$ does not present important variations in function of the projection angle $\theta$ with maximal difference recorded at $\theta\in\{\pi/4, 3\pi/4\}$. About the function support property, the SSRT-based basis function has compact support since the pixel basis function $\varphi$ is bounded. 

In this work, the utilization of such basis function in CT is motivated by the advantages provided by the SSRT such as good computation simplicity and accurate modeling of the CT system components.   

\section{SSRT Properties}
\subsection{Basic Properties}
In order to better handle the equations dealing with the SSRT properties, let us define the unitary vectors in the $(r-s)$ coordinates system, as  $\bm{\xi}=[\cos\theta\ \sin\theta]^t$ and $\bm{\xi}^{\bot}=[-\sin\theta\ \cos\theta]^t$, the position vector as $\bm{z}=[x \ y]^t$ and $d\bm{z}=dxdy$. Thus, $(\Delta)$ equation, in Fig.~\ref{ssrt_def}, may be written as $\rho=\bm{\xi}^t\bm{z}=x\cos\theta+y\sin\theta$ and then the SSRT can be rewritten as    
\begin{equation}
\Re_{\sigma}f(\rho,\bm{\xi})=\int f(\bm{z})g_\sigma(\bm{\xi}^t\bm{z}-\rho) d\bm{z}
\end{equation}

Several elementary properties of the SSRT of a function $f$ defined in $\mathbb{R}^2$ follow directly from the definition.  
 
\subsubsection{Linearity}
Consider 2 functions $f_1$ and $f_2$, 2 real numbers $a$ and $b$ and a function $h$ such as $h=af_1+bf_2$, we obtain 

\begin{align}\nonumber
\Re_{\sigma}h(\rho,\bm{\xi})&=
\int\left[af_1(\bm{z})+bf_2(\bm{z})\right]g_\sigma(\bm{\xi}^t\bm{z}-\rho) d\bm{z}\\
&=a\Re_{\sigma}f_1(\rho,\bm{\xi})+b\Re_{\sigma}f_2(\rho,\bm{\xi})
\label{lnrty}
\end{align}  

\subsubsection{Homogeneity}
If $a\in\mathbb{R}^*$, then we have

\begin{align}\nonumber
\Re_{\sigma}f(a\rho,a\bm{\xi})&=\int f(\bm{z})g_\sigma(a\bm{\xi}^t\bm{z}-a\rho) d\bm{z}\\\nonumber
&=\frac{1}{|a|}\int f(\bm{z})g_{\sigma/|a|}(\bm{\xi}^t\bm{z}-\rho) d\bm{z}\\
&=\frac{1}{|a|}\Re_{\sigma/|a|}f(\rho,\bm{\xi})
\label{homogenty}
\end{align}

Because of the presence of the amounts $\sigma/|a|$ in the result, the property of homogeneity is not satisfied for the SSRT, except when $|a|=1$, i.e. $a=\pm{1}$. For $a=1$, it consists in the identity transform whilst $a=-1$ relates the case of symmetry in regards to the axes system origin. The last property will be detailed in the next paragraph.

\subsubsection{Symmetry}
Setting $a=-1$ in (\ref{homogenty}) yields 
\begin{equation}
\Re_{\sigma}f(-\rho,-\bm{\xi})=\Re_{\sigma}f(\rho,\bm{\xi})
\end{equation}
In another notation, it means: $\Re_{\sigma}f(-\rho,\theta\pm\pi)=\Re_{\sigma}f(\rho,\theta)$.

\subsubsection{Geometric transformation}
Let $\bm{z}$ and $\bm{z'}$, two vectors representing two points in $\mathbb{R}^2$ space so that 
\begin{equation}
\bm{z'} = \bm{A}\bm{z}
\end{equation}
where $\bm{A}$ is a nonsingular matrix with real elements representing the transfer matrix of geometric transformations such as scaling, rotation, shearing, etc.
\begin{equation}
\Re_{\sigma}[f(\bm{z'})](\rho,\bm{\xi})=\int f(\bm{A}\bm{z})\cdotp g_\sigma(\bm{\xi}^t\bm{z}-\rho) d\bm{z}
\end{equation}
Since $\bm{z} = \bm{A}^{-1}\bm{z'}$ and $d\bm{z}=\left|\bm{A}^{-1}\right|d\bm{z'}$ where $|\cdotp|$ denotes the matrix determinant, then, we obtain
\begin{align}\nonumber
\Re_{\sigma}[f(\bm{z'})](\rho,\bm{\xi})&=\left|\bm{A}^{-1}\right|\int f(\bm{z'})\cdotp g_\sigma(\bm{\xi}^t\bm{A}^{-1}\bm{z'}-\rho) d\bm{z'}\\\nonumber
&=\left|\bm{A}^{-1}\right|\Re_{\sigma}[f(\bm{z})](\rho,(\bm{\xi}^t\bm{A}^{-1})^t)\\
&=\left|\bm{A}^{-1}\right|\Re_{\sigma}[f(\bm{z})](\rho,\bm{A}^{-t}\bm{\xi})
\end{align}

\paragraph{Case of Rotation:}
If $\bm{A}$ is a matrix of rotation, i.e. $\bm{A}^{t}=\bm{A}^{-1}$ and $\left|\bm{A}^{-1}\right|=1$, we have

\begin{equation}
\Re_{\sigma}[f(\bm{z'})](\rho,\bm{\xi})=\Re_{\sigma}[f(\bm{z})](\rho,\bm{A}^t\bm{\xi})
\end{equation}

In other words, if $\bm{A}=M_{Rot}^{\theta_0}=\begin{bmatrix}\cos\theta_0 & \sin\theta_0 \\-\sin\theta_0 & \cos\theta_0\end{bmatrix}$ then $\bm{A}^t\bm{\xi}=\begin{bmatrix}\cos(\theta+\theta_0) \\\sin(\theta+\theta_0)\end{bmatrix}$. In former notation, this means that

\begin{equation}
\Re_{\sigma}^{R_{\theta_0}}f(\rho,\theta)=\Re_{\sigma}f(\rho,\theta+\theta_0)
\end{equation}
 
\paragraph{Case of Scaling:}
Let $\bm{A}=\alpha\mathbb{I}$ where, $\alpha \ (\alpha\in\mathbb{R}_+^*)$ a scale factor, then $\bm{A}^{-1}=\frac{1}{\alpha}$ and $\left|\bm{A}^{-1}\right|=1/\alpha^2$. Hence, we obtain
\begin{equation}
\Re_{\sigma}[f(\bm{z'})](\rho,\bm{\xi})=\frac{1}{\alpha^2}\Re_{\sigma}[f(\bm{z})]\left(\rho,\frac{\bm{\xi}}{\alpha}\right)
\end{equation}  
On the other hand,

\begin{align}\nonumber
\Re_{\sigma}[f(\bm{z'})](\rho,\bm{\xi})
&=\frac{1}{\alpha^2}\int f(\bm{z'})g_\sigma\left(\frac{\bm{\xi}^t}{\alpha}\bm{z'}-\rho\right) d\bm{z'}\\\nonumber
&=\frac{1}{\alpha^2}\int f(\bm{z'})g_\sigma\left[\frac{1}{\alpha}\left(\bm{\xi}^t\bm{z'}-\alpha\rho\right)\right] d\bm{z'}\\\nonumber
&=\frac{1}{\alpha^2}\int f(\bm{z'})\left[\alpha g_{\alpha\sigma}\left(\bm{\xi}^t\bm{z'}-\alpha\rho\right)\right] d\bm{z'}\\
&=\frac{1}{\alpha}\Re_{\alpha\sigma}[f(\bm{z})]\left(\alpha\rho,\bm{\xi}\right)
\end{align} 

In terms of former notation, we have 
\begin{equation}
\Re_{\sigma}^{S_\alpha}f(\rho,\theta)=\frac{1}{\alpha}\Re_{\alpha\sigma}f(\alpha\rho,\theta)
\end{equation}

\subsubsection{Translation}
Let $\bm{z}$, $\bm{z'}$ and $\bm{b}$ three vectors representing the coordinates of 3 points in $\mathbb{R}^2$ space such as $\bm{z'}=\bm{z}+\bm{b}$. In order to find a relationship between the SSRTs of the functions $f(\bm{z})$ and $f(\bm{z'})$, it suffices to write
\begin{align}\nonumber
\Re_{\sigma}[f(\bm{z'})](\rho,\bm{\xi})
&=\int f(\bm{z}+\bm{b})g_\sigma\left(\bm{\xi}^t\bm{z}-\rho\right) d\bm{z}\\
&=\int f(\bm{z'})g_\sigma\left(\bm{\xi}^t\bm{z'}-\bm{\xi}^t\bm{b}-\rho\right) d\bm{z'}
\end{align} 
which yields  
\begin{equation}
\Re_{\sigma}[f(\bm{z'})](\rho,\bm{\xi})=\Re_{\sigma}[f(\bm{z})](\rho+\bm{\xi}^t\bm{b},\bm{\xi})
\end{equation}
Using another notation, if $\bm{b}=[b_x \ b_y]^t$ then the translation property can be expressed as
\begin{equation}
\Re_{\sigma}^{T_{\bm{b}}}f(\rho,\theta)=\Re_{\sigma}f(\rho+b_x\cos\theta+b_y\sin\theta,\theta)
\end{equation}

\subsection{Analytical Inversion}\label{inv_transf_subsect}
According to (\ref{ssrt_conv}), we can write 
\begin{equation}
\mathcal{R}f(\rho,\theta)=\mathcal{F}_\rho^{-1}\left[\frac{\mathcal{F}_\rho[\Re_{\sigma}f(\rho,\theta)](\omega)}{\mathcal{F}_\rho[g(\rho)](\omega)}\right](\rho,\theta)
\label{rad_ssrt_freq}
\end{equation}
where $\mathcal{F}_\rho$ and $\mathcal{F}_\rho^{-1}$ are, respectively, the 1D Fourier transform and its inverse with regards to the variable $\rho$.
Let us recall that the inverse Radon transform theorem states that

\begin{equation}
f(x,y)=\frac{1}{2\pi^2}\int_0^\pi\int_{-\infty}^{\infty}\frac{\partial \mathcal{R}f(\rho,\theta)}{\partial \rho}\frac{1}{x\cos\theta+y\sin\theta-\rho}d\rho d\theta  
\label{inv_rad}
\end{equation}
for $0\leq\theta<\pi$ and $-\infty<\rho<\infty$.

By substituting (\ref{rad_ssrt_freq}) in (\ref{inv_rad}), we obtain
\begin{equation}
f(x,y)=\frac{1}{2\pi^2}\int_0^\pi\int_{-\infty}^{\infty}\frac{\partial \mathcal{F}_\rho^{-1}\left[\frac{\mathcal{F}_\rho[\Re_{\sigma}f(\rho,\theta)](\omega)}{\mathcal{F}_\rho[g(\rho)](\omega)}\right](\rho,\theta)/ \partial \rho}{x\cos\theta+y\sin\theta-\rho}d\rho d\theta  
\label{inv_ssrt}
\end{equation}

\section{Image reconstruction using Scale Space Radon Transform}

\subsection{SSRT inversion through image deconvolution}\label{deconv_subsect}
The computation of the inverse SSRT (iSSRT) through (\ref{inv_ssrt}), evoked in \textsection\ref{inv_transf_subsect}, is straightforward but it is just given as an indication, since the deconvolution defined by the division in Fourier space is hazardous. 
On the basis of the link between the SSRT and the RT, provided in (\ref{ssrt_conv}), the inversion can be implemented in two steps:

\begin{itemize}
\item \textbf{Step 1}: Estimate the Radon transform $\mathcal{R}f()$ given the SSRT $\Re_{\sigma}f()$ and the Gaussian function $g()$ using an existing deconvolution algorithm.
\item \textbf{Step 2}: Estimate the image function $f()$ by using an existing RT inversion algorithm.
\end{itemize}

For Step 1, efficient deconvolution methods to estimate the Radon transform could be applied such as the image deconvolution method using the regularized filter algorithm \cite{Gonzalez02}. About Step 2., dealing with image reconstruction from RT projections, various approaches are investigated and developed by the researchers, among them, the popular Filtered Backprojection reconstruction method which is based on the Fourier Slice Theorem (FST). With the aim to compare the reconstruction methods in terms of performance, the deconvolution scheme in \cite{Gonzalez02}, followed by FBP, will be used later in the Experiments. 

In the following two subsections, we will go over FST and FBP, which are frequently utilized with RT, in order to address SSRT inversion. 

\subsection{Fourier Slice Theorem and Frequency-Domain Reconstruction for SSRT}
The 1-D Fourier transform (FT) of Radon transform w.r.t. to the variable $\rho$, noted $P_{\theta}(\omega)$ is given by
\begin{equation}
P_{\theta}(\omega)= \mathcal{F}_\rho(\mathcal{R}f(\rho,\theta))=\int_{-\infty}^{+\infty}\mathcal{R}f(\rho,\theta)e^{-i2\pi\omega\rho} d\rho
\label{fourier_1d}
\end{equation}
where, $\omega$ is the frequency.

By substituting the RT expression into (\ref{fourier_1d}), we obtain 
\begin{multline}
P_{\theta}(\omega)= \\
\int_{-\infty}^{+\infty}\left[\int\int_{-\infty}^{+\infty}f(x,y)\delta(x\cos\theta + y \sin\theta-\rho) dx dy\right]e^{-i2\pi\omega\rho} d\rho
\end{multline}
where, $\delta$ is the Dirac function.

Rearranging the order of integrals and using the sifting property of Dirac's delta function, gives

\begin{equation}
P_{\theta}(\omega)= \int\int_{-\infty}^{+\infty}f(x,y)e^{-i2\pi\omega(x\cos\theta + y \sin\theta)} dx dy
\end{equation}

Comparing now this equation with the expression of the 2-D FT of $f(x,y)$, given by

\begin{equation}
\mathcal{F}(u,v)= \int\int_{-\infty}^{+\infty}f(x,y)e^{-i2\pi(ux+vy)} dx dy
\end{equation}

We remark that the right side of the expression $P_{\theta}(\omega)$ represents the 2D FT of $f(x,y)$ expressed in polar coordinates system of the Fourier plan, which results in the proposed statement

\begin{equation}
P_{\theta}(\omega) =\mathcal{F}(u,v)|_{u=\omega\cos\theta,v=\omega\sin\theta}=\mathcal{F}_{polar}(\omega,\theta)
\label{SLF_propos}
\end{equation}

Thus, in parallel-beam geometry, the FST states that the 1-D Fourier transform $P_{\theta}(\omega)$ of a projection $\mathcal{R}f(\rho,\theta)$, obtained from the image $f(x,y)$, is identical to the central slice, at the angle $\theta$, of the 2-D image spectrum $\mathcal{F}(u,v)$ \cite{Jan06}.

Now, how could we apply the FST on the SSRT? For this purpose, for given $\theta$ and $\sigma$, let us compute the 1-D FT of the SSRT w.r.t. the variable $\rho$. Using (\ref{ssrt_conv}) and FT of the convolution we can write

\begin{align}\nonumber
P_{\theta,\sigma}(\omega)&= \mathcal{F}_\rho[\Re_{\sigma}f(\rho,\theta)](\omega)=\mathcal{F}_\rho[g(\rho)](\omega)\cdotp\mathcal{F}_\rho[\mathcal{R}f(\rho,\theta)](\omega)\\
&=G(\omega)P_{\theta}(\omega)
\label{ssrt_w_conv}
\end{align}

where, $G(\omega)=e^{-2\pi^2\sigma^2\omega^2}$ is the Fourier transform of the Gaussian function $g$. 

Since, FST deals with the RT projections, the spectrum of the latter, which is linked to the 2-D FT of $f(x,y)$ as seen above, can be deduced from the SSRT projections as 

\begin{equation}
P_{\theta}(\omega)=P_{\theta,\sigma}(\omega)/G(\omega)=P_{\theta,\sigma}(\omega)/e^{-2\pi^2\sigma^2\omega^2}
\label{ssrt_rt_w}
\end{equation}

We can see, for small values of $G(\omega)$ i.e., for very large values of $\sigma$ and/or $\omega$, that computing $P_{\theta}(\omega)$ by (\ref{ssrt_rt_w}) becomes problematic, since, divisions by a value close to zero will greatly increase high-frequency noise contribution.

To address this issue, among others, a Wiener filter could be used. Before going further, let us consider a discrete $M\times N$ image $f$, defined on the image domain $D_f$ and corrupted by additive white noise $\eta$ with zero mean and variance $\sigma_n^2$. For each pixel $(x,y)$, we define the observed image $f_n$ as $f_n(x,y) = f(x,y) + \eta(x,y)$. Using the linearity property in (\ref{lnrty}), the SSRT of the observed image $f_n$ is given by

\begin{align}\nonumber
\Re_{\sigma}f_n(\rho,\theta) &= \Re_{\sigma}f(\rho,\theta) + \Re_{\sigma}\eta(\rho,\theta)\\
&=(\mathcal{R}f\ast^\rho g)(\rho,\theta) + \Re_{\sigma}\eta(\rho,\theta)
\label{noisy_prj}
\end{align}

and, for a given $\theta$, we can compact (\ref{noisy_prj}) to obtain 

\begin{equation}
\Re_{\theta,\sigma}f_n(\rho) = (\mathcal{R}_{\theta}f\ast^\rho g)(\rho) + \Re_{\theta,\sigma}\eta(\rho)
\label{noisy_prj_comp}
\end{equation}

The formulae (\ref{noisy_prj_comp}) represents a typical mathematical model for many imaging systems including CT, in which $g$ is commonly known as the point spread function (PSF). This function being known, which consists, in our case, in Gaussian function, deconvolution operations permit to reverse the process in order to improve the resolution of the imaging system. In our case, the deconvolution consists in the estimation of the RT projection from that of SSRT in presence of noise and Gaussian smoothing. Numerous algorithms could be developed to recover $\mathcal{R}_{\theta}f$ from (\ref{noisy_prj_comp}). Examples include Wiener deconvolution \cite{Wiener49,Dhawan85}, Lucy-Richardson algorithm \cite{Richardson72} and variational model \cite{You96,Welk16}. Wiener deconvolution applies a Wiener filter inherent in the deconvolution to the observed noisy samples, enabling simultaneous reconstruction and denoising \cite{Dhawan85}. It is designed so that the expectation value of the Mean Square Error (MSE) $\epsilon=\mathbb{E}\{(\mathcal{R}_{\theta}f(\rho)-\hat{\mathcal{R}}_{\theta}f(\rho))^2\}$ is minimized where, the \textit{hat} symbol is used to indicate the estimate of the noise-free RT projection. By deriving the MSE minimization in the frequency domain, the 1-D FT, w.r.t. $\rho$, of the noise-free RT projection estimate is given by \cite{Gonzalez02} 

\begin{equation}
\hat{P}_{\theta}(\omega) = H_{W}(\omega)P^{f_n}_{\theta,\sigma}(\omega)
\label{SSRT_Rad_rel}
\end{equation}

where, $P^{f_n}_{\theta,\sigma}(\omega)$ is the SSRT projection spectrum of the observed noisy image and $H_{W}(\omega)$ is the frequency response of the Wiener filter, expressed as
\begin{equation}
H_{W}(\omega) = \frac{1}{G(\omega)}\frac{|G(\omega)|^2}{|G(\omega)|^2+K}
\label{Wiener_filt}
\end{equation}
where, $K=\frac{P_w^\eta(\omega)}{P_w(\omega)}$ with $P_w(\omega)=\mathbb{E}|P_\theta(\omega)|^2$ and $P_w^\eta(\omega)=\mathbb{E}|P_{\theta,\sigma}^\eta(\omega)|^2$ are the mean power spectral densities of $\mathcal{R}_{\theta}f(\rho)$ and $\Re_{\theta,\sigma}\eta(\rho)$, respectively. 

When there is zero noise, i.e. $K=0$, the Wiener filter is simply the inverse filter. In practice, the quantities $P_w^\eta(\omega)$ and $P_w(\omega)$ are not known; so $K$ is set to a constant scalar which is determined empirically. 
\begin{figure}[!tbh]
\centering
\includegraphics[scale=0.5]{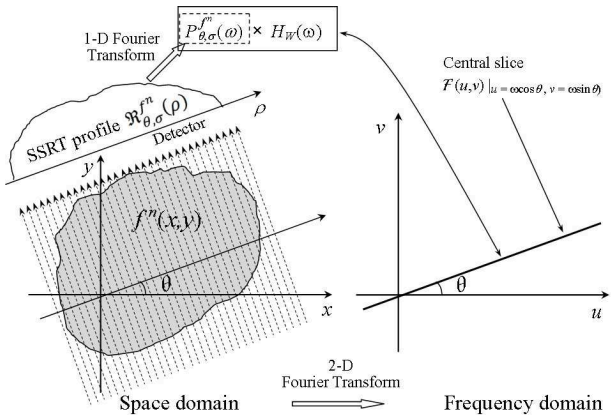}
\caption{Illustration of Fourier Slice Transform in case of SSRT} 
\label{ssrt_sft}
\end{figure}

In case of SSRT, the FST flowchart can be illustrated in Fig~\ref{ssrt_sft}. So, by rotating the SSRT detector for 180\degree, the entire 2D FT $\mathcal{F}(u,v)|_{u=\omega\cos\theta,v=\omega\sin\theta}$ is "measured". Once the 2D frequency space of the object function is filled, a 2-D inverse FT then yields the reconstructed object function, i.e.

\begin{equation}
f(x,y)= \int\int_{-\infty}^{+\infty}\mathcal{F}(u,v)e^{i2\pi(ux+vy)} du dv
\end{equation}

The method of image reconstruction, depicted until now, is called \textit{Frequency-Domain Reconstruction}. As ascertained in \cite{Kak88}, in practice, only a finite number of projections of an object can be taken. Then, the function $\mathcal{F}(u,v)$ is only known along a finite number of radial lines. To overcome this limitation and fill completely the 2D frequency space of the object function, very time-consuming operations of interpolation are required. Furthermore, since the density of the radial points becomes sparser as one gets farther away from the center, the interpolation error also becomes larger. These errors, implied in the high frequency domain, result in image degradation \cite{Kak88}. Consequently, the Frequency-Domain Reconstruction method is not used in practice.

\subsection{Filtered Backprojection (FBP) using SSRT for Parallel-Beam Geometry}\label{FBP_subsect}
As with RT, the FST will be applied on the SSRT projections to derive the FBP algorithm. This derivation is based on the reformulation of the FST into polar coordinates, yielding a 2-step reconstruction method: a projection filtering and a backprojection onto the image domain. 

The 2D inverse FT of $\mathcal{F}(u,v)$ is given by

\begin{equation}
f(x,y)= \mathcal{F}^{-1}\{\mathcal{F}(u,v)\}=\int\int_{-\infty}^{+\infty}\mathcal{F}(u,v)e^{i2\pi(ux+vy)} du dv
\end{equation}

Expressing this equation in a polar coordinate system $(\omega,\theta)$ with $u=\omega\cos\theta,v=\omega\sin\theta$ and $dudv=\omega d\omega d\theta$, yields

\begin{equation}
f(x,y) = \int_{0}^{2\pi}\int_{0}^{+\infty}\mathcal{F}_{polar}(\omega,\theta)e^{i2\pi\omega(x\cos\theta+y\sin\theta)} \omega d\omega d\theta
\end{equation}

Splitting the outer integral domain into $[0,\pi)$ and $[\pi,2\pi)$, and using $\mathcal{F}_{polar}(\omega,\theta+\pi)= \mathcal{F}_{polar}(-\omega,\theta)$ yields

\begin{equation}
f(x,y) = \int_{0}^{\pi}\left[\int_{-\infty}^{+\infty}\mathcal{F}_{polar}(\omega,\theta)|\omega| e^{i2\pi\omega(x\cos\theta+y\sin\theta)} d\omega\right] d\theta
\end{equation}

According to the FST proposal (\ref{SLF_propos}) and the RT/SSRT relationship in frequency domain provided in (\ref{SSRT_Rad_rel}), we obtain
  
\begin{equation}
f(x,y) = \int_{0}^{\pi}\left[\int_{-\infty}^{+\infty}P^{f_n}_{\theta,\sigma}(\omega)H_{W}(\omega)|\omega|e^{i2\pi\omega(x\cos\theta+y\sin\theta)}  d\omega\right] d\theta
\label{FBP_eq}
\end{equation}

The integral in (\ref{FBP_eq}) may be expressed as 

\begin{equation}
f(x,y) = \int_{0}^{\pi} Q_{\theta,\sigma}(x\cos\theta+y\sin\theta)d\theta
\label{backprj_operat}
\end{equation}
 where

\begin{equation}
Q_{\theta,\sigma}(\rho)=\int_{-\infty}^{+\infty}P^{f_n}_{\theta,\sigma}(\omega)H_{W}(\omega)|\omega|e^{i2\pi\omega\rho}  d\omega\\
\label{filtering_operat}
\end{equation}

The formulae (\ref{filtering_operat}) represents a filtering operation, where the frequency response of the filter is given by $H_s(\omega)=|\omega|H_{W}(\omega)$ whilst, (\ref{backprj_operat}) represents the backprojection of the filtered projections. The resulting projections for the different angles $\theta$ are stacked to form the estimate of $f(x,y)$ which represents the reconstructed image. Let recall, as seen in literature, that the \textit{ramp} filter, with frequency response $H(\omega)=|\omega|$, is retrieved in the FBP reconstruction algorithm, used for RT inversion. However, according to the present founding, it is important to note that, in case of SSRT, the FBP procedure leads to a convolution of the projections with a filter kernel $h_s$ of which the frequency response is equal to the product $H_{s}(\omega)=H(\omega)H_{W}(\omega)$. According to the reasoning developed until now, we have considered the idealized case where, there is a continuum of projection views. In practice, since the image function is discrete and bounded, the sinograms have only a finite number of angular and radial samples. Then, the formula (\ref{filtering_operat}) and (\ref{backprj_operat}) dealing with the FBP reconstruction method using SSRT require modification for practical implementations. In this work, the discrete approximations used in \cite{Kak88} and \cite{Fessler09} are the basis of the reconstruction algorithm, depicted above.

As $H_{s}(\omega)$ is not absolutely integrable, it is not possible to derive directly the corresponding impulse response by the inverse FT. To address this issue, we have to window the filter so that it becomes zero outside a defined frequency interval.

Suppose that $\mathcal{R}_{\theta}f(\rho)$ is band-limited with maximum frequency less than $\omega_{max}=\frac{1}{2\tau}$ where, $\tau$ is the sampling interval with which the projection data are measured. Then,

\begin{equation}
    H_s(\omega)= 
\begin{cases}
    |\omega|H_W(\omega),&  |\omega|<\omega_{max}\\
    0,              & \text{otherwise}.
\end{cases}
\label{filt_freq_resp}
\end{equation}

\begin{figure}[!tbh]
\centering
\includegraphics[scale=0.4]{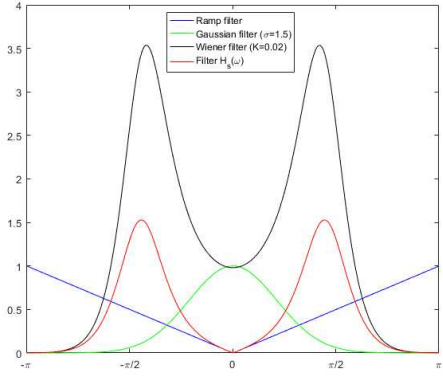}
\caption{Example of the filter response $H_s(\omega)$ (in red line) for the FBP reconstruction using SSRT with $\sigma=1.5$ and $K=0.02$} 
\label{filt_resp}
\end{figure}

An example of frequency response for $H_s(\omega)$ is given in Fig.~\ref{filt_resp} where, the constant $K$ and the Gaussian kernel $std$ $\sigma$, for Wiener filter, are set to 0.02 and 1.5, respectively.

It is known that the ramp filter $H(\omega)=|\omega|$ amplifies high frequency noise, so the low-pass Wiener filter $H_W$ is used, here, as a smoothing filter. 

The steps of the FBP algorithm applied on SSRT projections are summarized as follows.

\begin{itemize}
\item \textbf{Step 1}: Compute the 1-D Fourier transform $P^{f_n}_{\theta,\sigma}(\omega)$ of each SSRT projection for the observed image.
\item \textbf{Step 2}: Multiply each SSRT projection spectrum by the Wiener filter $H_W(\omega)$ to obtain the spectrum estimate $\hat{P}_\theta(\omega)$ of the noise-free RT projection, as in (\ref{SSRT_Rad_rel}).
\item \textbf{Step 3}: Multiply $\hat{P}_\theta(\omega)$, for each $\theta$, by the 1-D windowed ramp filter $H(\omega)=|\omega|$
\item \textbf{Step 4}: Perform a 1-D inverse FT on the filtered projection using (\ref{filtering_operat}).
\item \textbf{Step 5}: Perform the backprojection by integrating, over $\theta$, all the 1-D inverse FTs from Step 4 using (\ref{backprj_operat}).
\end{itemize}

\section{Experiments}

\begin{figure}
    \centering
    \includegraphics[scale=0.2]{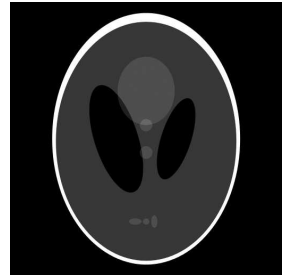} 
    \caption{Shepp-Logan phantom image}
    \label{Sh_Log_phant}
\end{figure}

The proposed FBP algorithm using SSRT projections, of which the steps are summarized in \textsection\ref{FBP_subsect}, is denoted SSRT-FBP. In order to show the effectiveness of this method in CT image reconstruction, let us compare its performance to: (1) the SSRT inversion scheme given at \textsection\ref{deconv_subsect}, denoted Deconv-Rad-FBP and (2) the classical FBP-based inverse Radon transform, denoted Radon-FBP. Because of its widespread utilization in the evoked domain, a Shepp-Logan phantom of size $M\times N=512\times 512$, coded on 256 gray levels, is used as test image in the first experimental analysis (see Fig.~\ref{Sh_Log_phant}). 

In order to compare the image reconstruction performance for each of the methods used, the influence of three factors are depicted: (1) the number of CT projections, (2) the value of the Gaussian kernel standard deviation $\sigma$ and (3) the type and the level of CT data noise.  

For the first factor, the test image is transformed by SSRT and RT to provide sinograms on $180\degree$, used as input projections, with angular steps of $\Delta\theta=0.5\degree, 1\degree, 2\degree, 5\degree, 10\degree, 20\degree, 30\degree$. 

As seen, previously, the SSRT projections magnitude depends on the standard deviation $\sigma$ of the Gaussian function involved in SSRT. That is why, to see the influence of the second factor on the outcomes, a range of $\sigma$ values, varying from $\sim 0$ to $\sigma=2.8$ with a step of $\Delta\sigma=0.2$, is employed in these first experiments.

If $S$ denotes the sinogram data $\mathcal{R}f(\rho,\theta)$ or $\Re_{\sigma}f(\rho,\theta)$ of an input object $f$, with $(\rho,\theta)$ as sinogram coordinates, produced, respectively, by the RT-based or SSRT-based CT scans, then, the one-dimensional (1-D) version of $S$, denoted $S^{1}$, will represent the sinogram as a vector with elements $S^1_i$, $i=1,\dots, N_S$, where $N_S$ is the resolution of the sinogram $S$ \cite{Xie17}. This resolution is nothing but the number of measurements in the scan which can be calculated as $N_S = m_S\times n_S$, where $m_S$ and $n_S$ are the number of discrete values of $\rho$ and $\theta$, representing the number of detector elements and the number of projection views, respectively.  

Using Lambert-Beer's law under the assumption of a monoenergetic X-ray beam, the mean of the transmission or the projection data can be expressed as $\bar{Y_i} = I_{0i}\exp\{-S^1_i\}$, where $I_{0i}$ is the incident flux representing the amount of photons along the projection path acquired by the detector bin $i$.

In practice, the CT data are noisy and the measurements are often modeled as a sum of a Poisson distribution representing photon-counting statistics and an independant Gaussian distribution with mean $m_n$ and variance $\sigma^2_n$ representing additive electronic noise \cite{Ding18,LaRiviere06,Singer13,Fu17}, where $m_n$ is assumed to be zero because the data acquisition system (DAS) offsets can be measured by a dark scan and subtracted from the CT data, which does not modify the variance of the data \cite{Fu17}.
The quanta noise is due to the limited number of X-ray photons collected by the detector and the electronic noise is the result of electronic fluctuation in the detector photodiode and other electronic components \cite{Ma12}. 
That is said, by taking into account the noise, the CT measurements are expressed as $Z_i = Y_i + \eta_i$ where, (1) $Y_i\sim Poisson(\lambda)$ is the quanta of rays received by detector bin $i$ with $\lambda=\bar{Y_i}$ is the mean number of photons or the noise-free transmission datum for detector bin $i$ at a projection view; and (2) $\eta_i\sim\mathcal{N}(0,\sigma_n^2)$, with $\sigma_n^2$ denoting the variance of the electronic noise which has been converted to photons for detector bin $i$. 

In these first experiments, to compare the performance of the used image reconstruction methods in presence of Poisson-Gaussian noise, we generated noisy projections of Shepp-Logan phantom with two incident exposure levels $I_0=\{10^5, 2\times 10^4\}$ photons per ray and two values for Gaussian noise variance $\sigma_n^2=\{0.01,1\}$.  

The reconstructed images by the methods Radon-FBP, Deconv-Rad-FBP and SSRT-FBP are denoted by the 2D functions $f^{iRad}$, $f^{deconv+iRad}$ and $f^{iSSRT}$, respectively. 
Well-known metric of Peak Signal-to-Noise Ratio (PSNR) and  Structural SIMilarity (SSIM) are used to evaluate the quality of the above-mentioned image reconstruction methods. Here, $PSNR_{k}=10\log_{10}\frac{255^2}{MSE_k}$ where,  $MSE_{k}=\sum_{i=1}^{M}\sum_{j=1}^{N}(f_{ij}-f_{ij}^k)^2\big/MN$ with, MSE is the acronym of the mean square error, $f$ is the free-noise input test image and $k\in\{iRad,deconv+iRad,iSSRT\}$. The higher the PSNR, the better the image reconstruction method; while the values of SSIM \cite{Wang11} belong to the range $[0\ 1]$ where, high values express better results.

To assure the statistical validity of the experiments since the noise is generated randomly, the calculations are carried out $N_r$ times ($N_r=20$) and then, the result is averaged. For the Wiener filter, a $K$ value of 0.02 proves to be sufficient to guarantee a trade-off between sharpening and smoothing effects. 
For all methods, the runtime is expressed in \textit{seconds} and the implementations are performed in Matlab environment on Dell Workstation with 2.9 GHz CPU and 32 GB RAM.

\begin{figure}
\centering
    \subfloat[Without noise]{\includegraphics[scale=0.33]{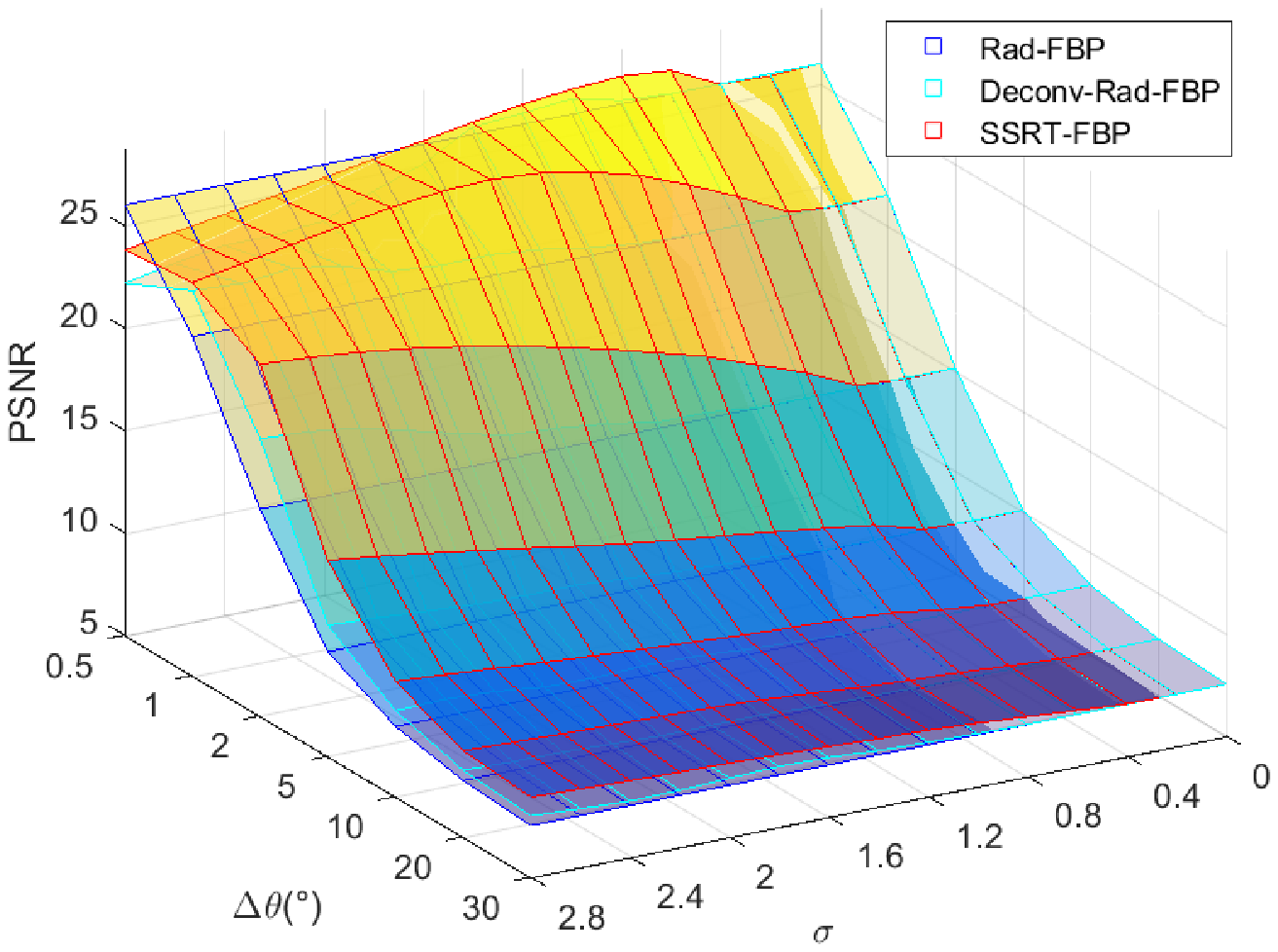}}\hspace*{-1.5em}
\subfloat[Without noise]{\includegraphics[scale=0.33]{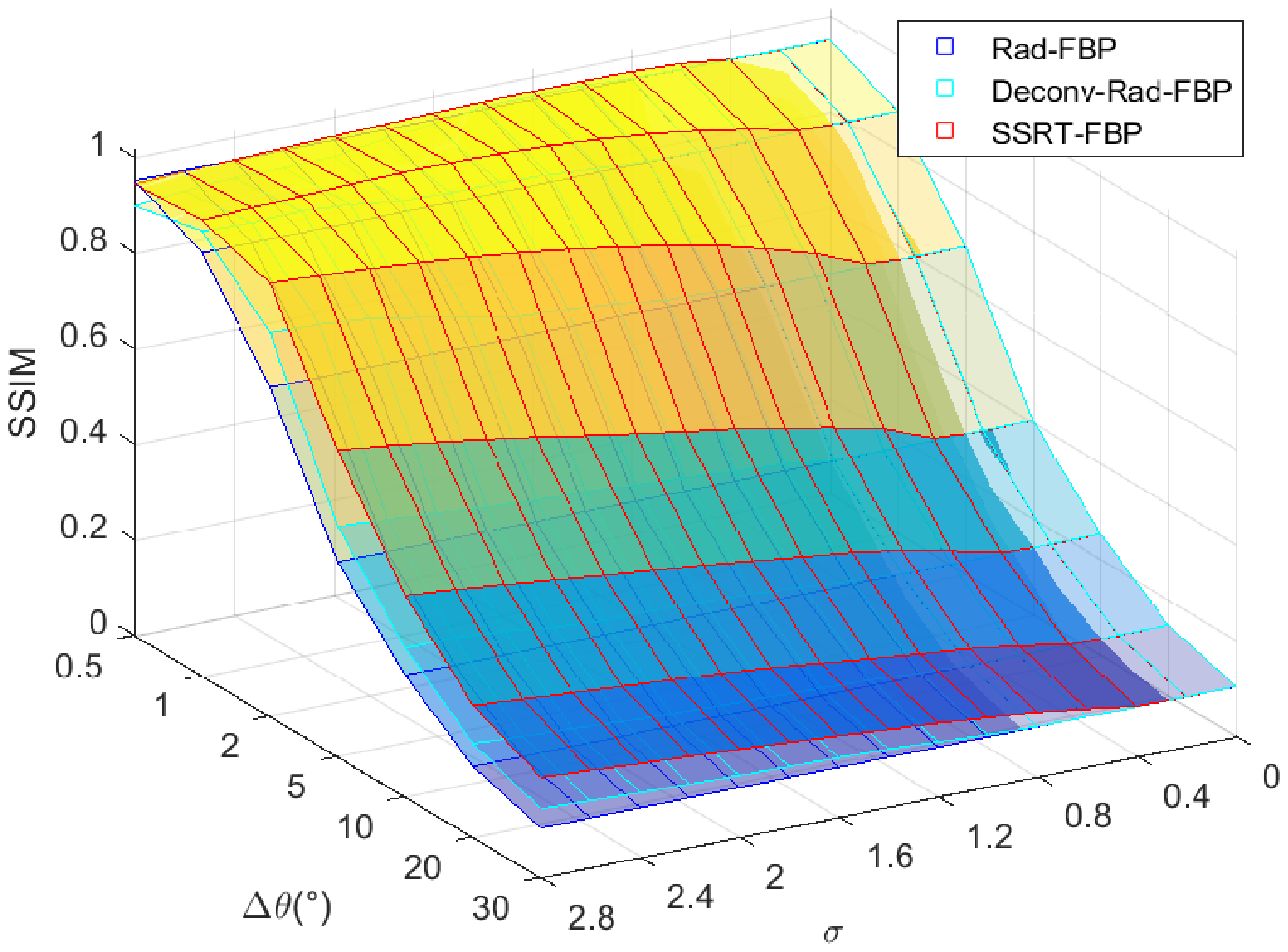}}\\
    \subfloat[$I_0=10^5,\sigma_n=0.1$]{\includegraphics[scale=0.33]{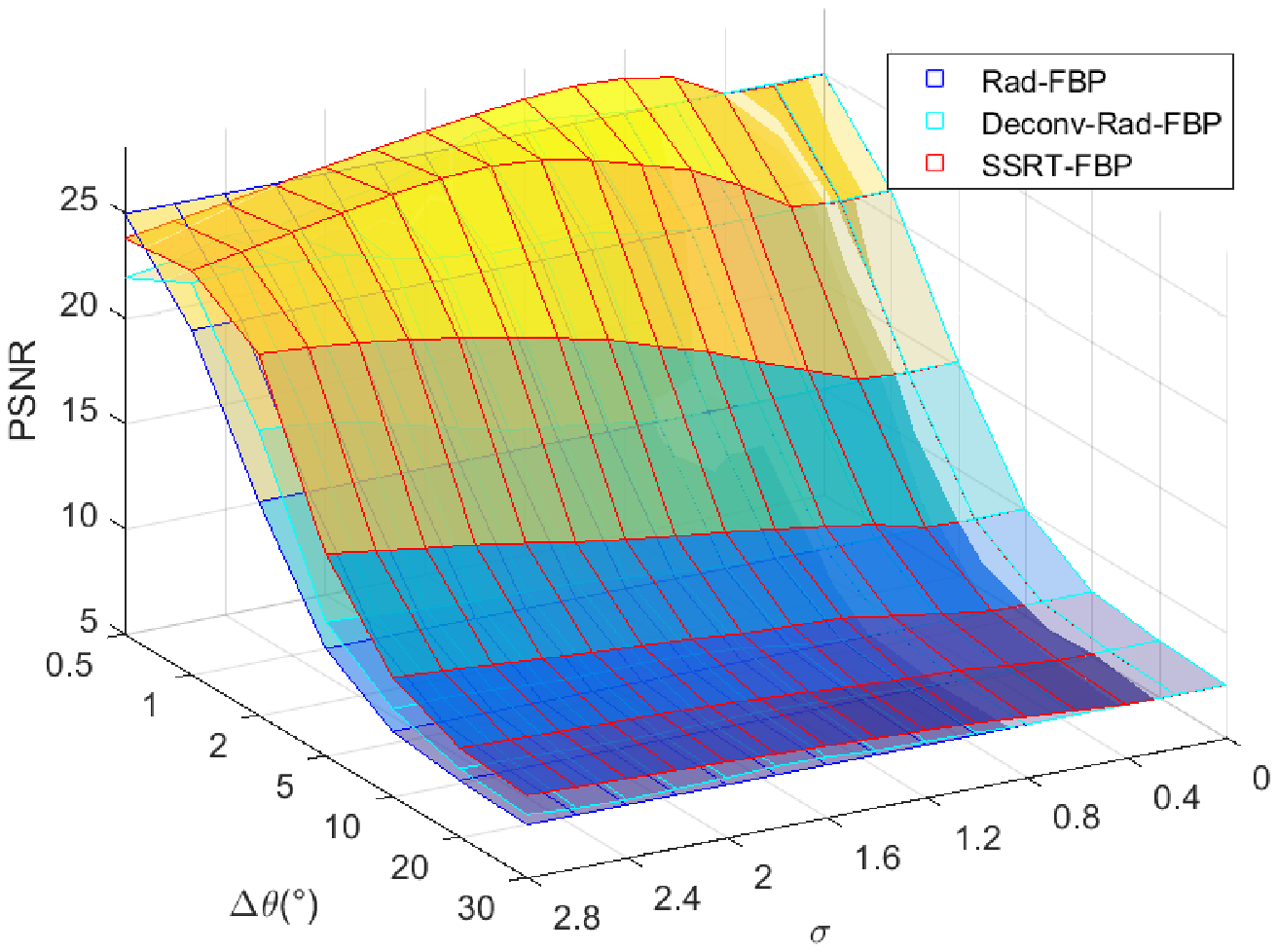}}\hspace*{-1.5em}
\subfloat[$I_0=10^5,\sigma_n=0.1$]{\includegraphics[scale=0.33]{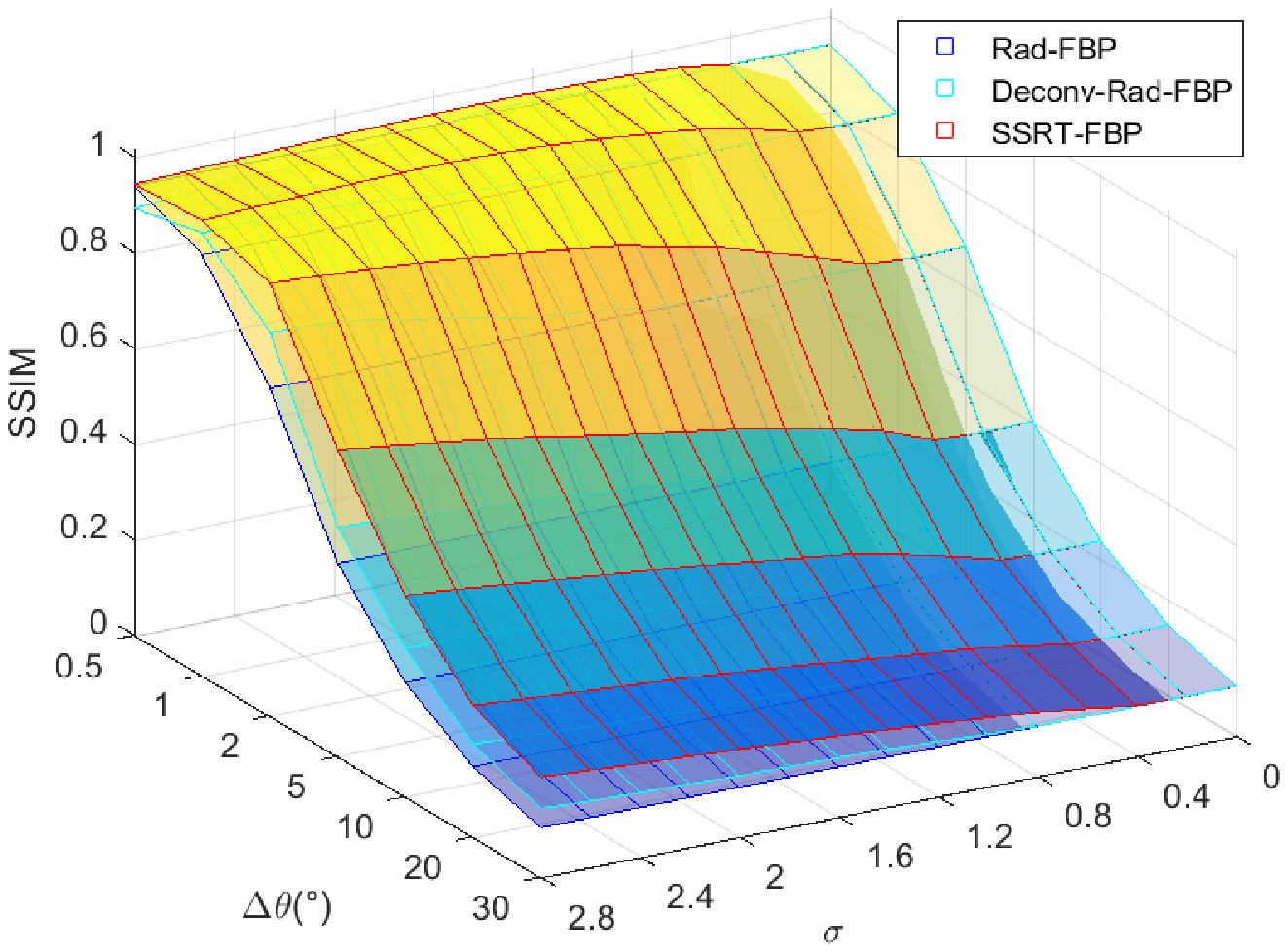}}\\
    \subfloat[$I_0=10^5,\sigma_n=1$]{\includegraphics[scale=0.33]{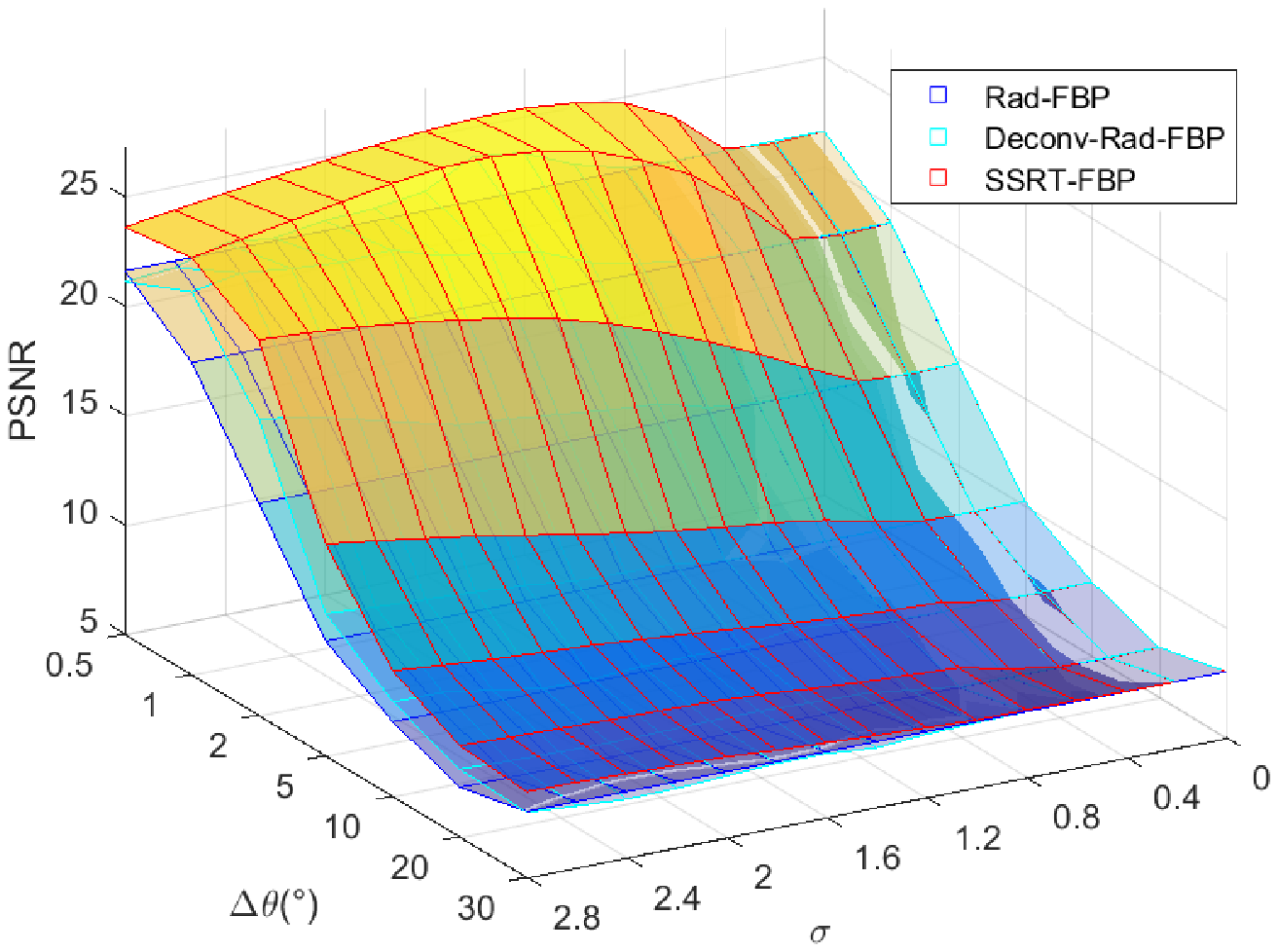}}\hspace*{-1.5em}
 \subfloat[$I_0=10^5,\sigma_n=1$]{\includegraphics[scale=0.33]{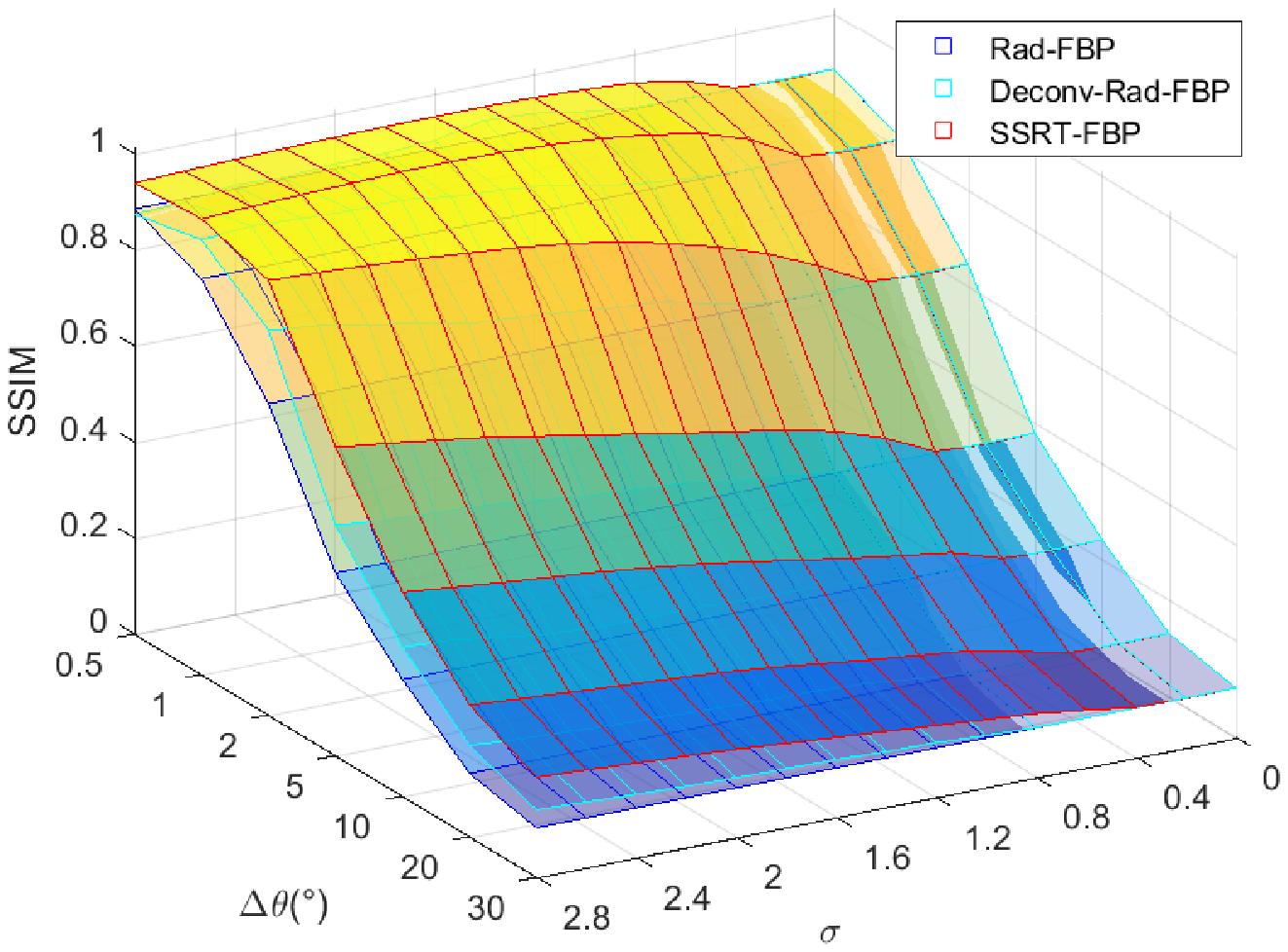}}\\
    \subfloat[$I_0=2\times 10^4,\sigma_n=0.1$]{\includegraphics[scale=0.33]{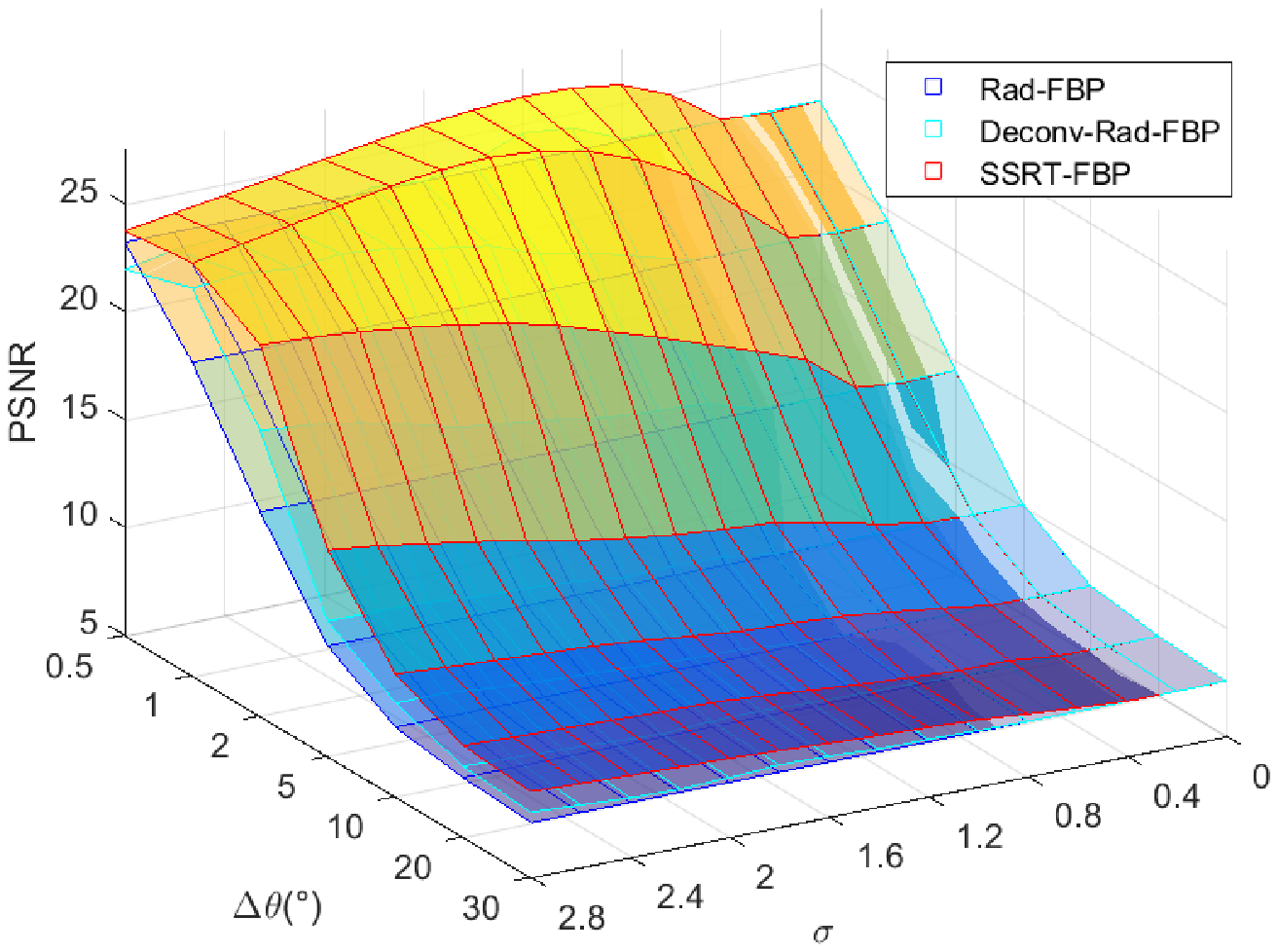}}\hspace*{-1.5em} 
 \subfloat[$I_0=2\times 10^4,\sigma_n=0.1$]{\includegraphics[scale=0.33]{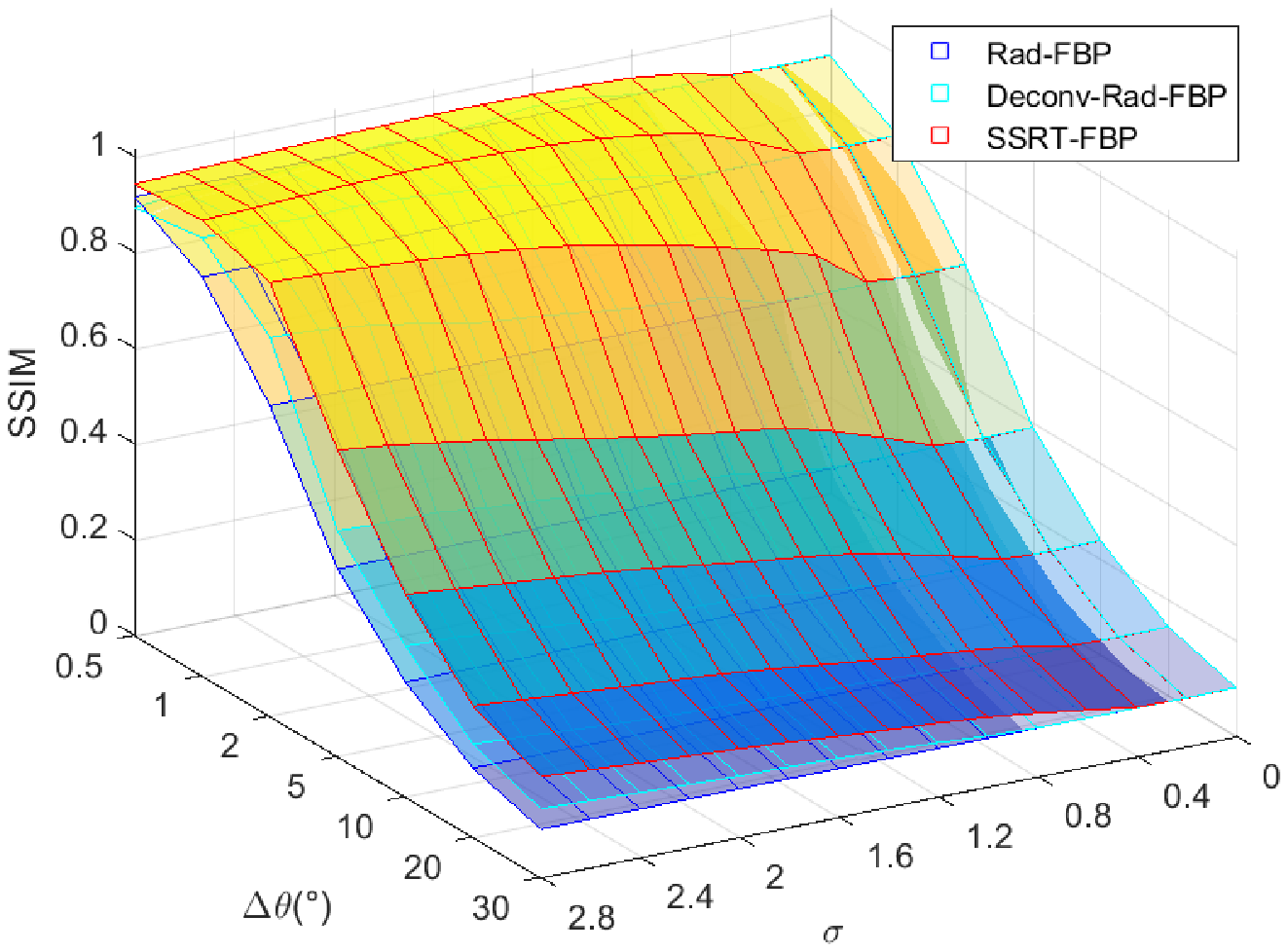}}\\ 
    \subfloat[$I_0=2\times 10^4,\sigma_n=1$]{\includegraphics[scale=0.33]{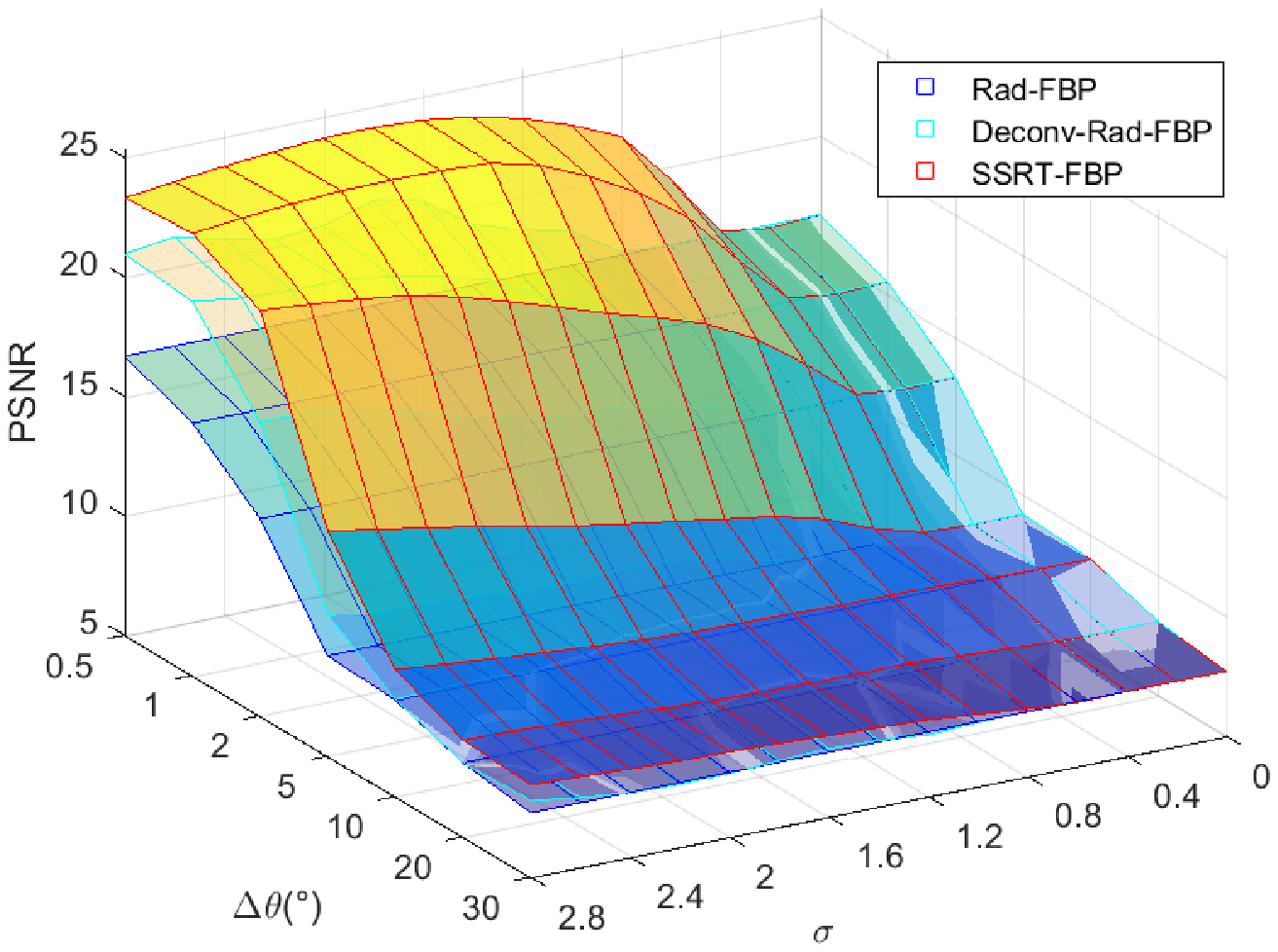}}\hspace*{-1.5em}
\subfloat[$I_0=2\times 10^4,\sigma_n=1$]{\includegraphics[scale=0.33]{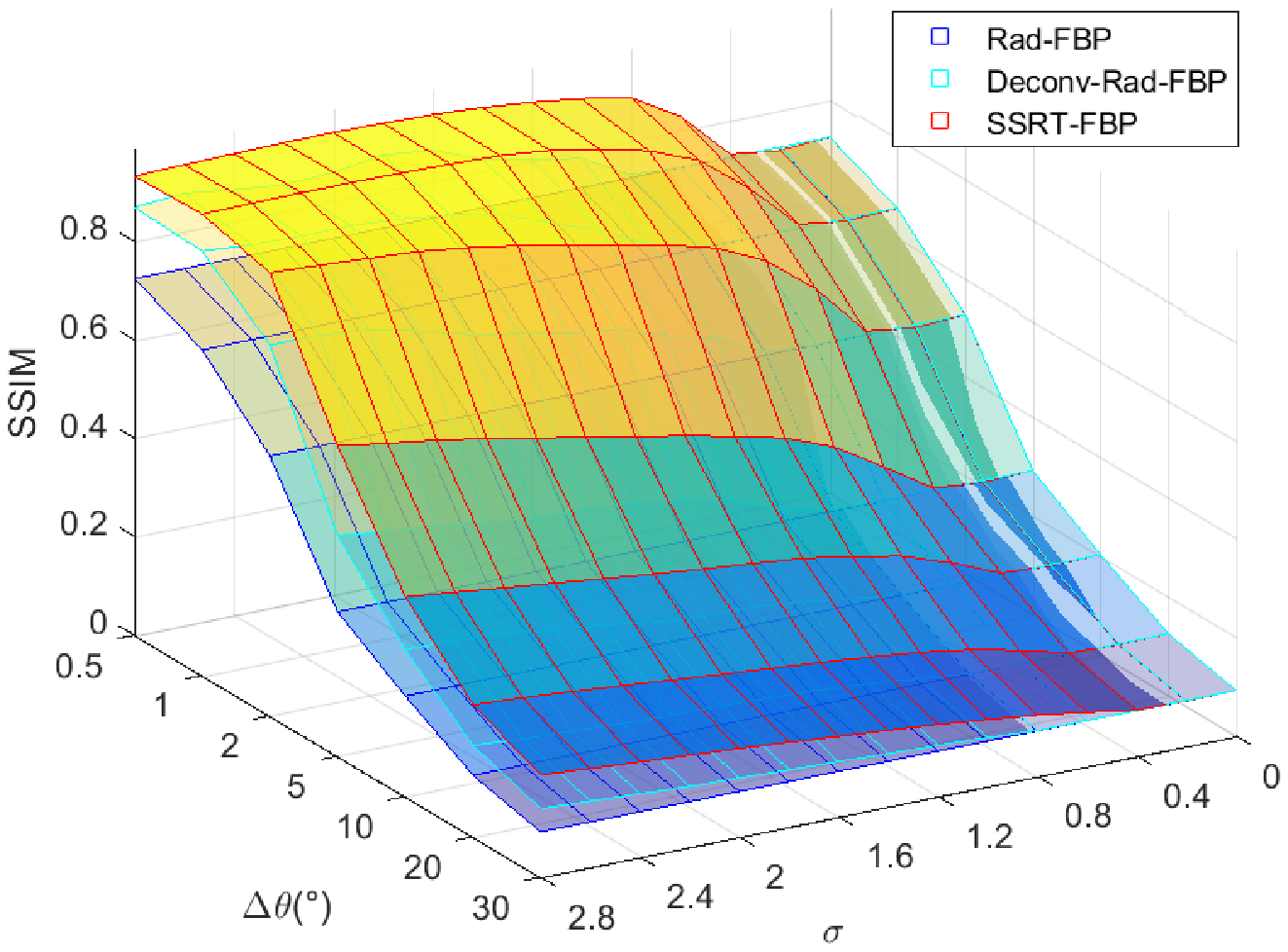}}\\
    \caption{PSNR and SSIM values for the reconstructed Shepp-Logan head phantom image in function of $\sigma$ and $\Delta\theta$ and noise parameters.}
    \label{psnr_ssim_shepp_log}
\end{figure}

Fig.~\ref{psnr_ssim_shepp_log} depict the PSNR and the SSIM values for Shepp-Logan image reconstruction by browsing $\sigma$ and $\Delta\theta$ on their value interval when the CT data noise is: (1) absent and (2) modeled by Poisson/Gaussian distributions with  above-mentioned noise levels. So, the interpretation and analysis of the PSNR and SSIM curves provided in Fig.~\ref{psnr_ssim_shepp_log}  will depend on the variables: (1) the SSRT scale parameter $\sigma$, (2) the number of projection which is linked with $\Delta\theta \left(\# \mathrm{projections} =180/\Delta\theta(\degree)\right)$ and (3) the levels of CT data noise which is, in turn, expressed as a mixture of : (3a) Poisson noise, parameterized by the number of incident photons $I_0$ and (3b) Gaussian white noise, parameterized by zero-mean electronic noise with variance $\sigma_n^2$. It is worth to note that, although the Radon-FBP method is not concerned by $\sigma$, the latter is just included in its PSNR and SSIM graphics for the sake of comparison. 

At first, let examine the PSNR and SSIM curves, provided in Figs~\ref{psnr_ssim_shepp_log}a and \ref{psnr_ssim_shepp_log}b, in case of free-noise CT data. It appears, clearly, that the image reconstruction, for all methods, is better as $\Delta\theta$ decreases; which is not surprising, since the quality of the reconstructed input Shepp-Logan phantom is increasingly improved as the number of projections increases. We can remark by running the $\sigma$ values, for each of the angle steps $\Delta\theta$, that the values of $\sigma$ less than $0.4-0.5$ give almost the same performance for all the methods.   
This is expectable since, for small $\sigma$'s, the behavior of SSRT-based methods tends to that of RT, as given in (\ref{ssrt_def_eq}). However, for greater $\sigma$ values, the SSRT-FBP method outperforms Radon-FBP and Deconv-Rad-FBP methods. As illustrated in Figs.~\ref{psnr_ssim_shepp_log}c-\ref{psnr_ssim_shepp_log}j, these observations are also valid when the input CT data is corrupted by Poisson-Gaussian noise with various levels where, the difference between the PSNR/SSIM curves, for the three methods, in the $\sigma-\Delta\theta$ space, is more pronounced. In fact, we observe that the PSNR/SSIM values dropped, drastically, for Radon-FBP and, to a lesser extent, for Deconv-Rad-FBP, as the noise level increases, i.e. when $I_0$ decreases and $\sigma_n$ increases. Inversely, the SSRT-FBP method is shown to be robust to the noise inherent to CT data. 
As first observation, the image reconstruction scores of the SSRT-based methods (Deconv-Rad-FBP and SSRT-FBP) are better than those obtained by Radon-FBP method where, particularly, the SSRT-FBP method is the best. The claimed outstanding performance is reasonable given that, in one hand, the SSRT-based approaches are more resistant to CT data noise thanks to the inclusion of a Gaussian kernel in their structural design which acts as a noise filter and, in the other hand, the SSRT-FBP method assumes that the X-ray beam projections are collected by detectors with a certain width, which match better with SSRT than with RT.
In fact, as explained in Section II and shown in Fig.~\ref{ssrt_def}, the projection made by the SSRT-based CT beam at $\rho$ along the line $(\Delta)$ actually realizes a Gaussian weighting sum of the gray levels of the pixels generating an image band with $(\Delta)$ as centerline. The thickness of this band which depends on the $\sigma$ value (see \cite{Ziou21}) is, in reality, the thickness of the X-ray beam projection on the CT detector. That is said, for RT-based CT, the displacement of the point $(0,\rho)$ in the $r-s$ plan is so small when the projection axis is rotated by smaller angles, i.e. $\Delta\theta\sim 0.5\degree-0.7\degree$, that practically all of the projections displayed on the RT sinogram are produced by all the pixels forming the input CT data. However, more noise and artifacts are added to the RT sinogram when the rotation angle increment is sufficiently large, i.e. when low-dose CT is used, because the missing projections result in interpolation errors on the reconstructed image space. 
As previously mentioned, the SSRT, on the other hand, manages a non-zero line projection which has a filtering role that enables it to cover the input image within number of projections. This SSRT characteristic could be advantageous for a variety of CT applications, such as medical CT, where the radiation dose reduction is essential since an excessive patient exposure to penetrating radiation results in major health harm. 

To better examine, for the tested methods, the variations of PSNR and SSIM in terms of variables $\Delta\theta$ and $\sigma$, let us focus in Fig.~\ref{psnr_ssim_shepp_log} on the sub-figures given at the first and last rows, corresponding to noise-free and the more noisy cases, respectively. In case of noise-free CT data (see Fig.~\ref{psnr_ssim_shepp_log}a and \ref{psnr_ssim_shepp_log}b), it can be seen that, for $\Delta\theta = 0.5\degree$, higher values of PSNR and SSIM are obtained by SSRT-FBP for $\sigma\sim 0.8$ whereas, for $\Delta\theta=1\degree$ and $\Delta\theta=2\degree$, the best results are still given by the last method when $\sigma\sim 1.6$ and $\sigma> 2.8$, respectively. The same remarks on the increase of $\sigma$ values could be made for larger $\Delta\theta$ values of over $2\degree$. In presence of pronounced noise as is the case in Fig.~\ref{psnr_ssim_shepp_log}i and \ref{psnr_ssim_shepp_log}j, highest quality reconstructions are obtained with SSRT-FBP when specific values of $(\Delta\theta,\sigma)$ pairs are taken such as $(\Delta\theta,\sigma)=\{(0.5\degree,1.8), (1\degree,2), (2\degree,>2.8)\}$ and the same observations as in noise-free case can be made for larger $\Delta\theta$. We remark that, in presence of noise, the $\sigma$ values, for the same values of $\Delta\theta$, are a bit greater than those in noise-free case. That is can be explained by the fact that $\sigma$, should be enough large, in presence of noise, so that the SSRT-FBP can achieve the noise filtering task, in addition of its role in minimizing the effects of small CT projections number.        

To support this explanation, we can observe, in noise-free case, that the Radon-FBP PSNR is slightly higher than the Deconv-Rad-FBP and SSRT-FBP PSNRs when $\Delta\theta$ is in the range $[0.5\degree \ 0.7\degree]$ and $\sigma$ is higher than 1.8 as shown, in Fig.~\ref{psnr_ssim_shepp_log}a. Due to the line integral principal, this observation could be justifiable since  the smaller $\Delta\theta$'s are, the bigger the accuracy of RT-FBP is in CT image reconstruction. So, in this case, above 1.8, $\sigma$ increases the smoothing effect of SSRT-FBP at the expense of the details preservation. However, for the same $\Delta\theta$, when the input CT data is corrupted by a strong composite noise, the RT-FBP performance deteriorates substantially, as illustrated in Fig.~\ref{psnr_ssim_shepp_log}i and \ref{psnr_ssim_shepp_log}j, which shows the usefulness of the proposed SSRT-FBP method in CT image reconstruction. 

The abovementioned observations ascertain the good behavior of SSRT-FBP in X-ray CT image reconstruction by handling correctly: (1) the missing data due to the reduction of the number of projections and (2) the presence of Poisson-Gaussian in CT sinograms. Consequently, the best recovery results could be obtained by a fine tuning of the Gaussian kernel standard deviation $\sigma$ in function of the number of body X-ray exposures.

\begin{figure}[tbh]
\centering
\subfloat[]{\includegraphics[scale=0.23]{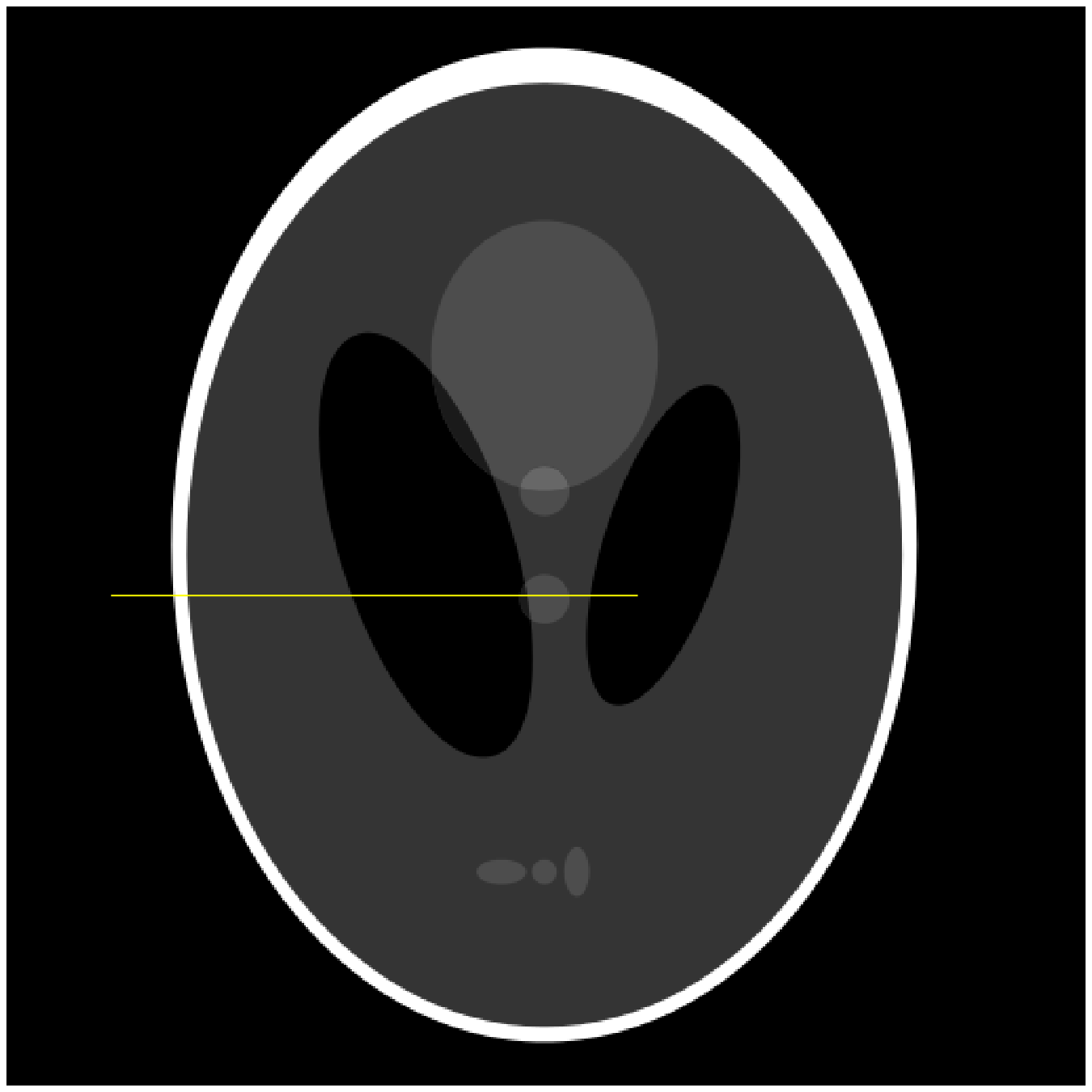}}
\subfloat[]{\includegraphics[scale=0.23]{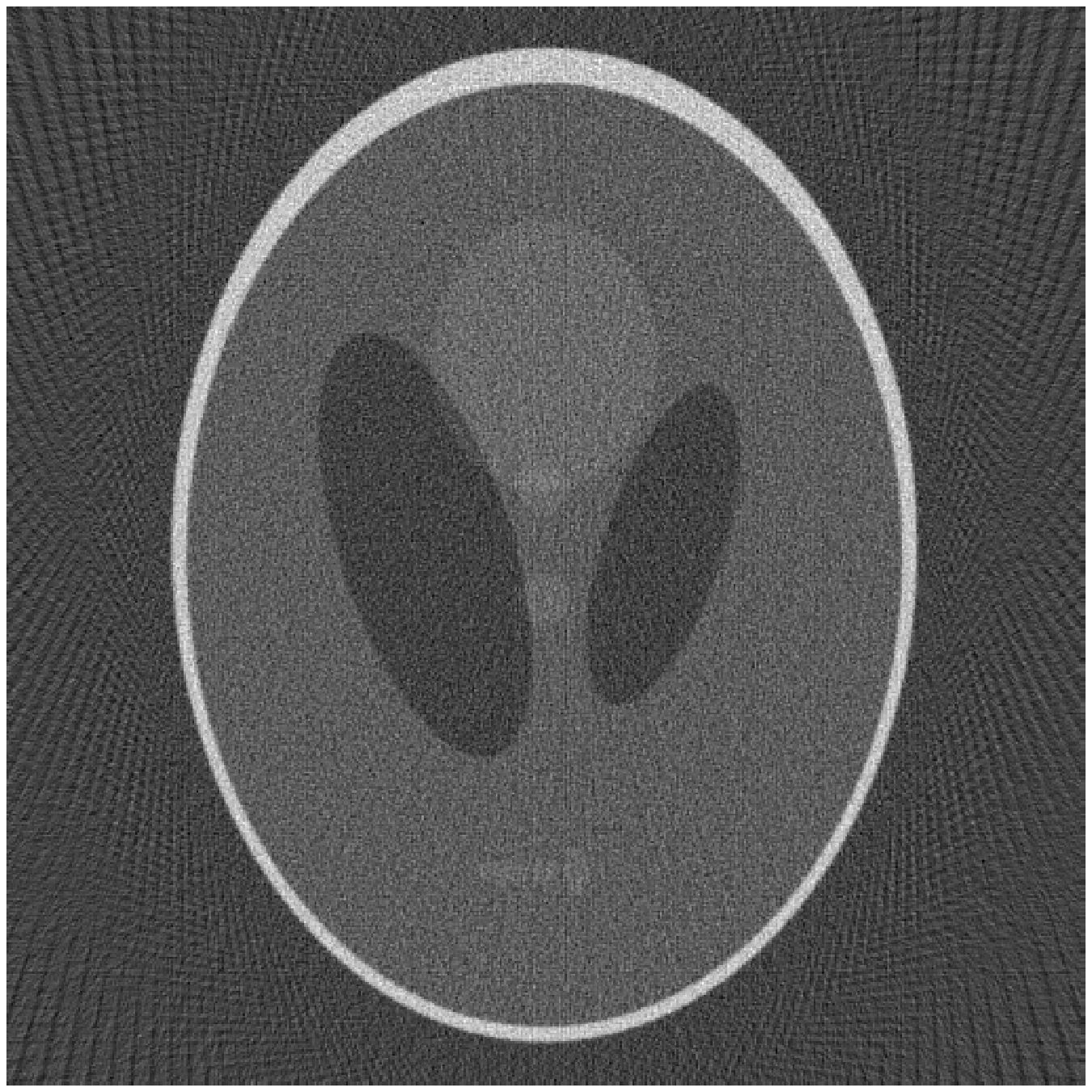}}\\ 
\subfloat[]{\includegraphics[scale=0.23]{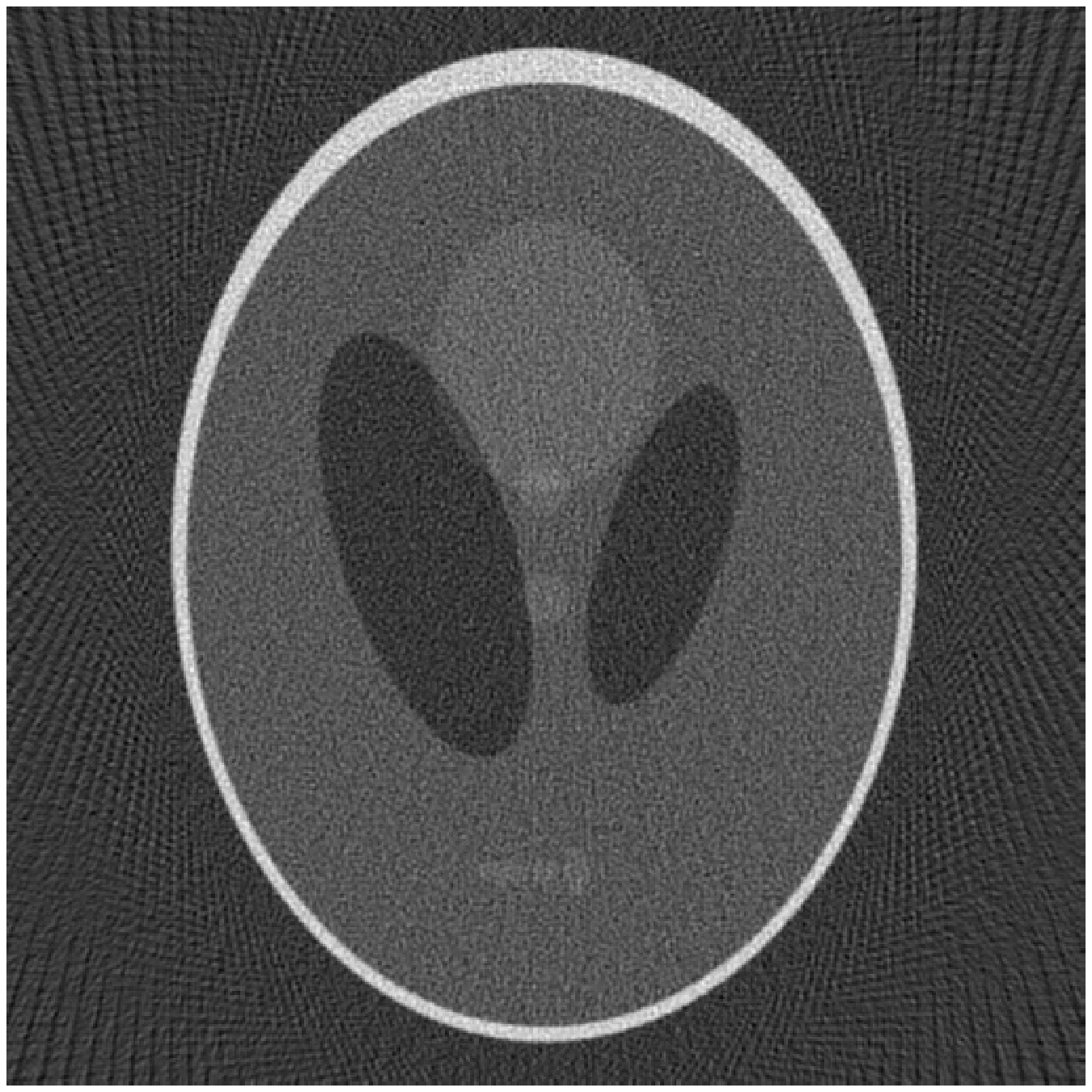}}
\subfloat[]{\includegraphics[scale=0.23]{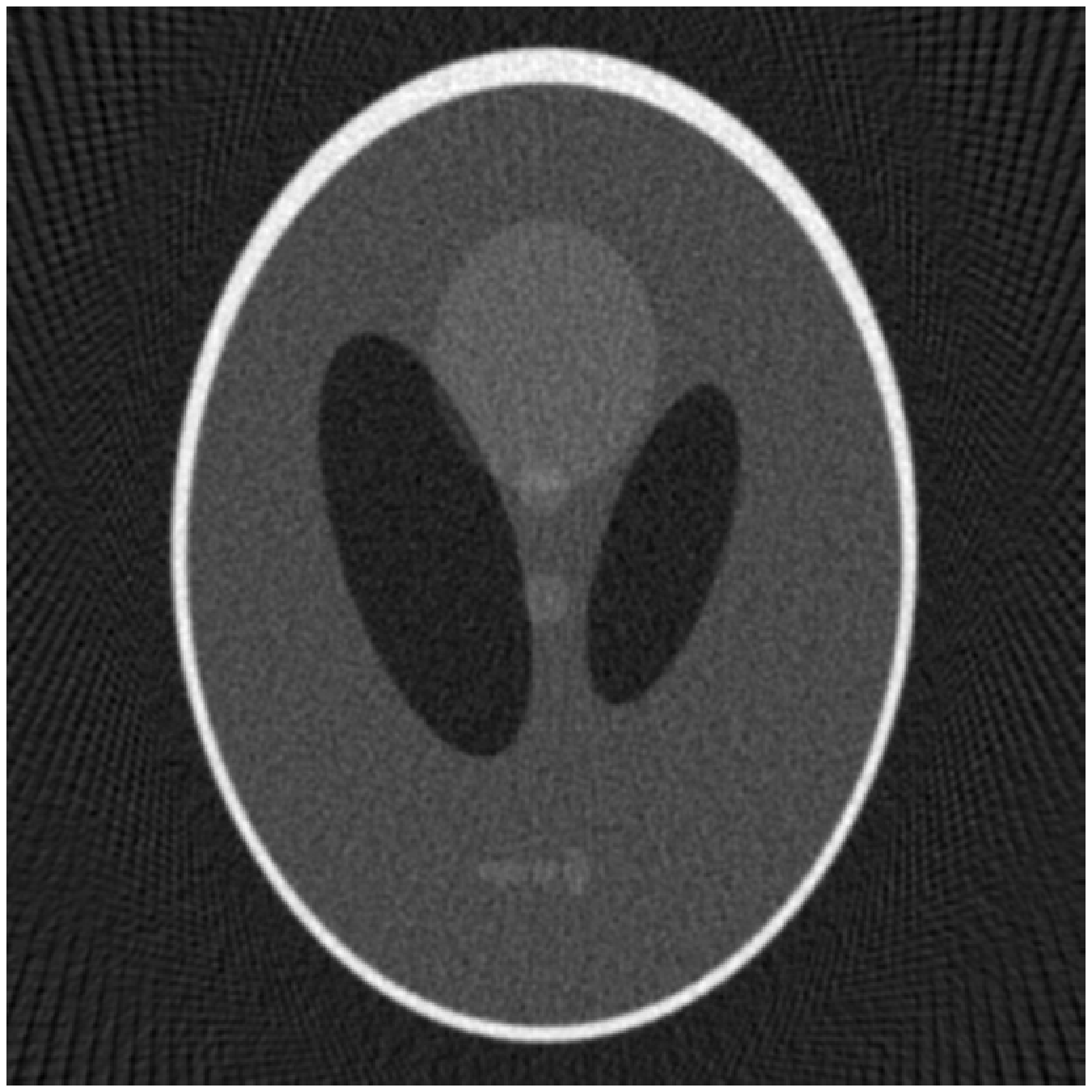}} 
\caption{(a) Unnoisy Shepp-Logan phantom for $(\sigma,\Delta\theta)=(1.2,2\degree)$. (b)-(d) Corrupted Shepp-Logan sinograms ($I_0=10^4$ and $\sigma_n=0.5$) and image reconstruction results by Radon-FBP, Deconv-Rad-FBP and SSRT-FBP, respectively.}
    \label{reconstr_im_res}
\end{figure}

Fig.~\ref{reconstr_im_res}a displays the noise-free image of Shepp-Logan phantom and Figs~\ref{reconstr_im_res}b-\ref{reconstr_im_res}d depicts the reconstructed images with methods of Radon-FBP, Deconv-Rad-FBP and SSRT-FBP, respectively, where $(\sigma,\Delta\theta)=(1.2,2\degree)$ and the Poisson-Gauss noise has as parameters $I_0=10^4$ and $\sigma_n=0.5$. 

\begin{figure}[tbh]
    \centering
    \includegraphics[scale=0.55]{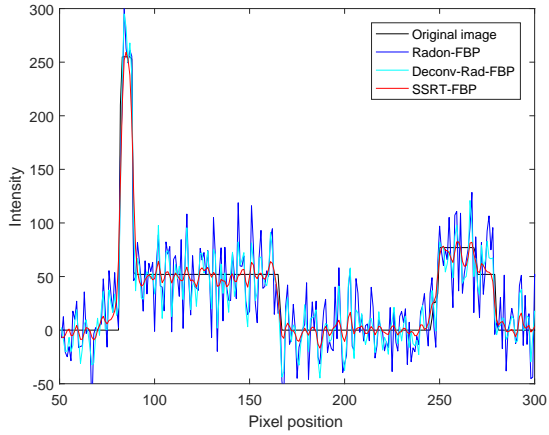}
    \caption{Intensity profiles from reconstructed images of noise-free and noisy Shepp-Logan CT data given in Fig.~\ref{reconstr_im_res} corresponding to yellow line segment drawn on Fig.~\ref{reconstr_im_res}a.}
    \label{profile_recons_im}
\end{figure}

The gray level profiles of the noise-free image and the abovementioned methods-based recovery results are depicted in Fig.~\ref{profile_recons_im}. The gray level profiles are located, in the $x-y$ plan, at the pixel positions $y=280$ and $x$ from 50 to 300, as illustrated by the yellow line segment in Fig.~\ref{reconstr_im_res}a.
It can be observed that the profiles of the results from SSRT-FBP are closer to that of the ground truth image than those of Deconv-Rad-FBP and Radon-FBP methods, where the fitting mean absolute errors of the profiles in the reconstructed images, compared to those of the ground truth, are 22.34, 15.14 and 7.35 for Radon-FBP, Deconv-Rad-FBP and SSRT-FBP, respectively. 
These last findings ascertain again the high noise reduction capability of SSRT-FBP, in addition of its ability to minimize the projections number reduction effects, as previously explained. 

The SSRT-FBP method has a runtime of about 0.03 s, which is equivalent to Radon-FBP in terms of execution speed, as shown in Table~\ref{table_runtime}. In fact, the SSRT-FBP routine in Matlab, called $issrt$, consists in adapting $iradon$ so that the SSRT scale parameter $\sigma$ is introduced in the various functions composing $iradon$ program and where, $Wiener$ filter with specific constant $K$ is used in $issrt$ while, only $ramp$ filter, designated by $ram-lak$, is used in $iradon$. So, both routines $iradon$ and $issrt$ follow the same algorithmic complexity. However, Deconv-Rad-FBP method takes around 0.09 s where, the time execution difference, compared to Radon-FBP, is spent on the deconvolution operation. 

\begin{table}[tbh]
\renewcommand{\arraystretch}{1.3}
\caption{Shepp-Logan image reconstruction runtime scores}
\label{table_runtime} 
\centering
\begin{tabular}{c|c c |c c|c c} 
&\multicolumn{2}{c|}{Radon-FBP}&\multicolumn{2}{c|}{Deconv-Rad-FBP}&\multicolumn{2}{c}{SSRT-FBP} \\
[2pt]\hline
Runtime (s) & \multicolumn{2}{c|}{0.029} & \multicolumn{2}{c|}{0.092} & \multicolumn{2}{c}{0.031}\\
\hline
\end{tabular}
\end{table}

In order to evaluate the effectiveness of the suggested reconstruction approaches in relation to the noise level and the number of projections, an anthropomorphic abdominal phantom was utilized in this work. The experimental phantom data were acquired from the work in \cite{Li16} where, the irradiated specimen consists in an anthropomorphic abdominal phantom with a removable
liver insert, designed by the research team and custom manufactured by QRM (Moehrendorf, Germany).  

Two liver inserts, each containing 19 embedded synthetic lesions with known volumes of varying diameter (6-40 mm), shape (spherical, ellipsoidal and lobulated), contrast  and density (homogenous and mixed) were designed to have liver parenchyma and lesion CT values simulating arterial phase (AP phantom) and portal venous phase (PVP phantom) imaging, respectively. 
The schematic of the phantom are illustrated by Fig. 1 in \cite{Li16}. 

\begin{figure}[tbh]
    \centering
    \subfloat[Abdominal phantom CT slice]{\includegraphics[scale=0.25]{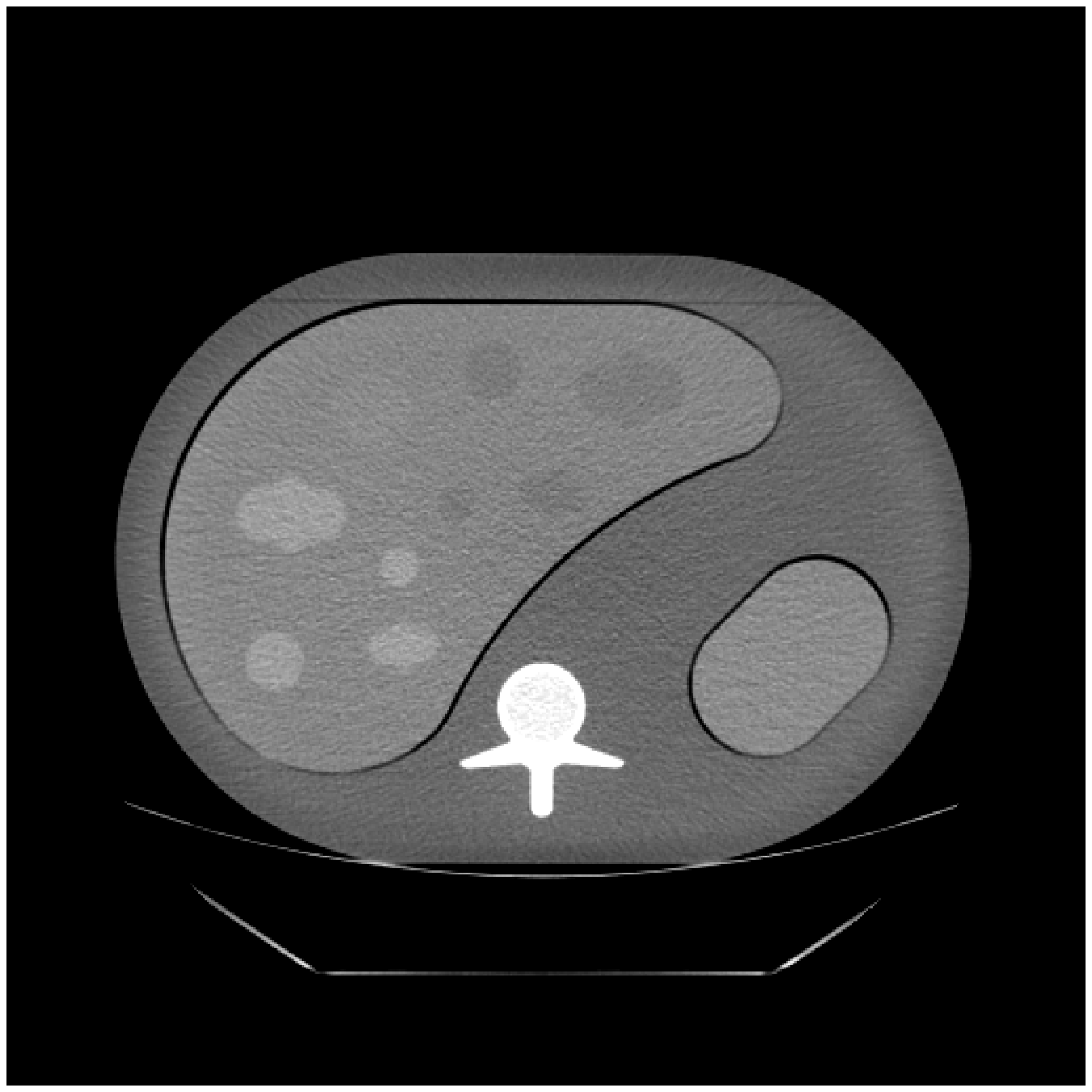}}\\
    \subfloat[Without noise]{\includegraphics[scale=0.33] {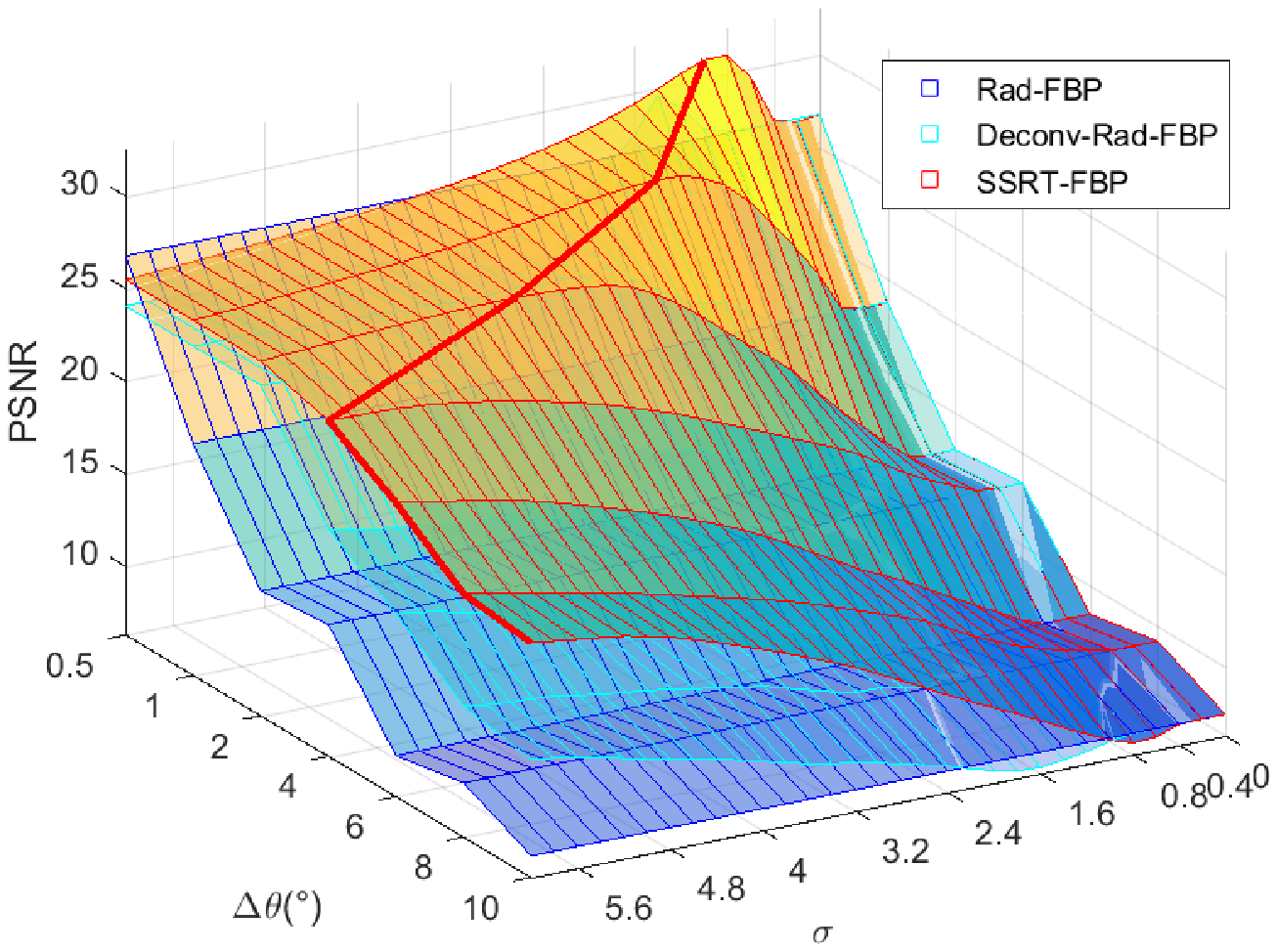}}  
             \hspace*{-1.5em}
    \subfloat[Without noise]{\includegraphics[scale=0.33]{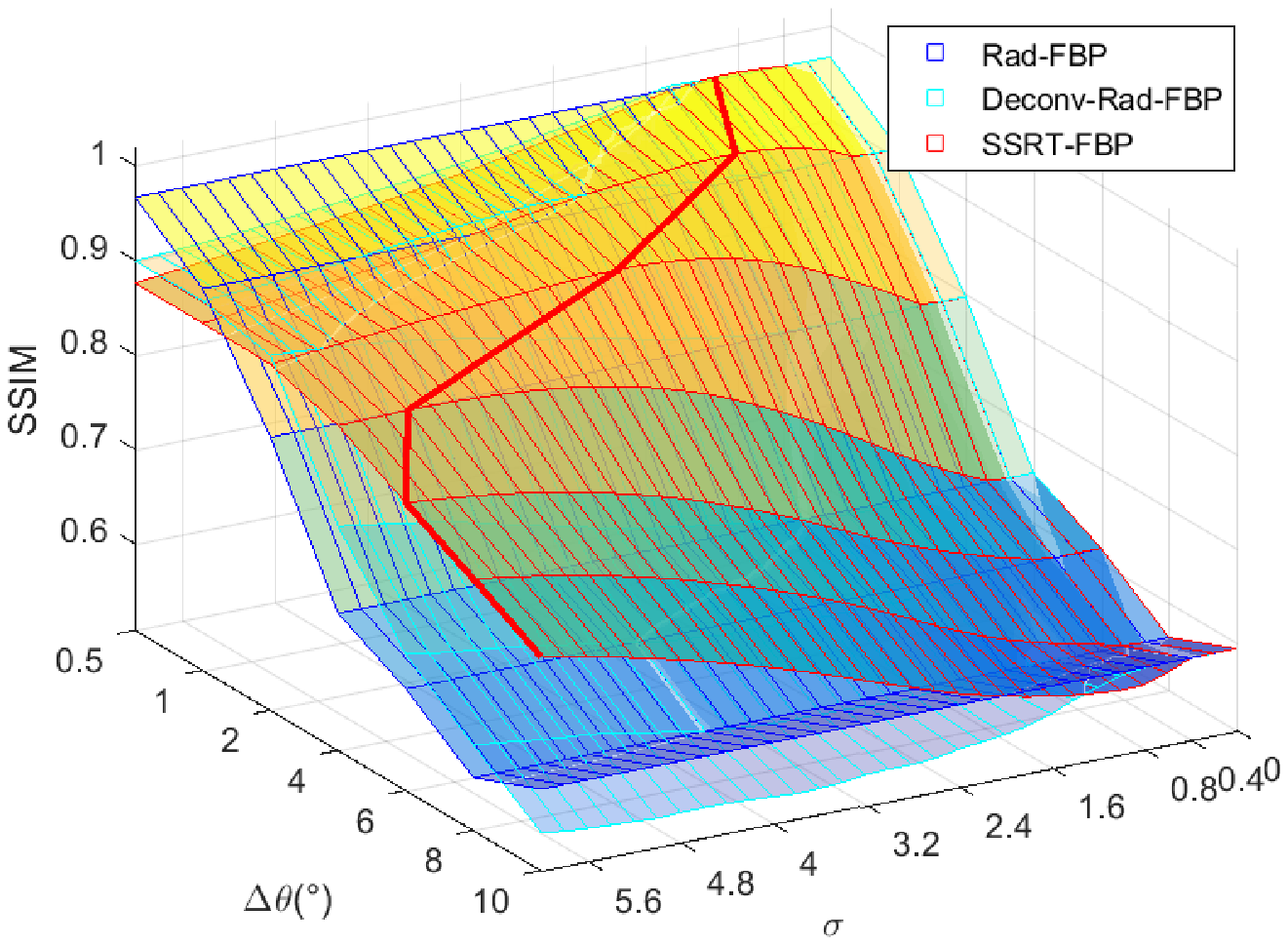}}\\  
    \subfloat[$I_0=5\times 10^4, \sigma_n=0.5$]{\includegraphics[scale=0.33]{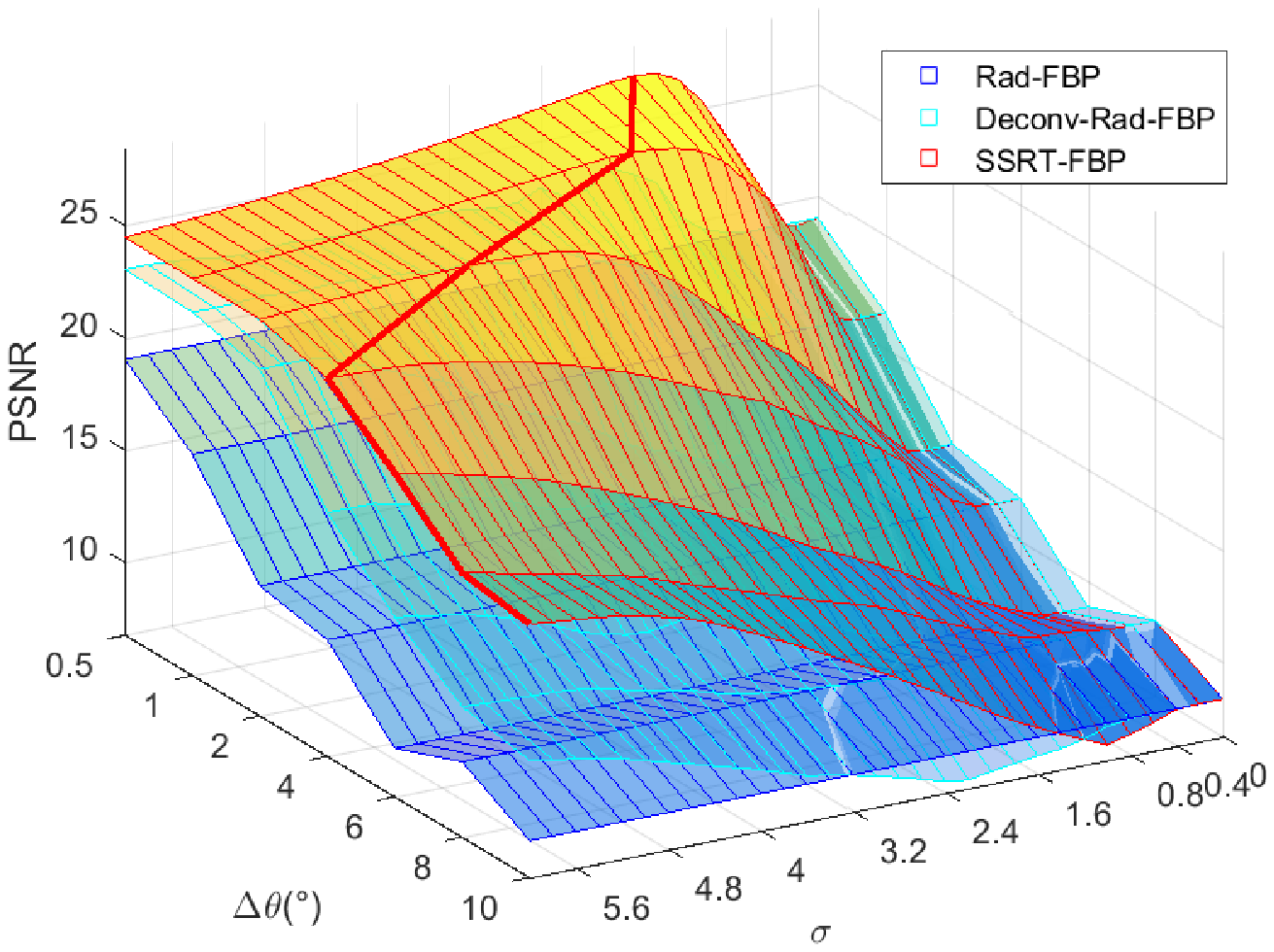}}\hspace*{-1.5em}
    \subfloat[$I_0=5\times 10^4, \sigma_n=0.5$]{\includegraphics[scale=0.33]{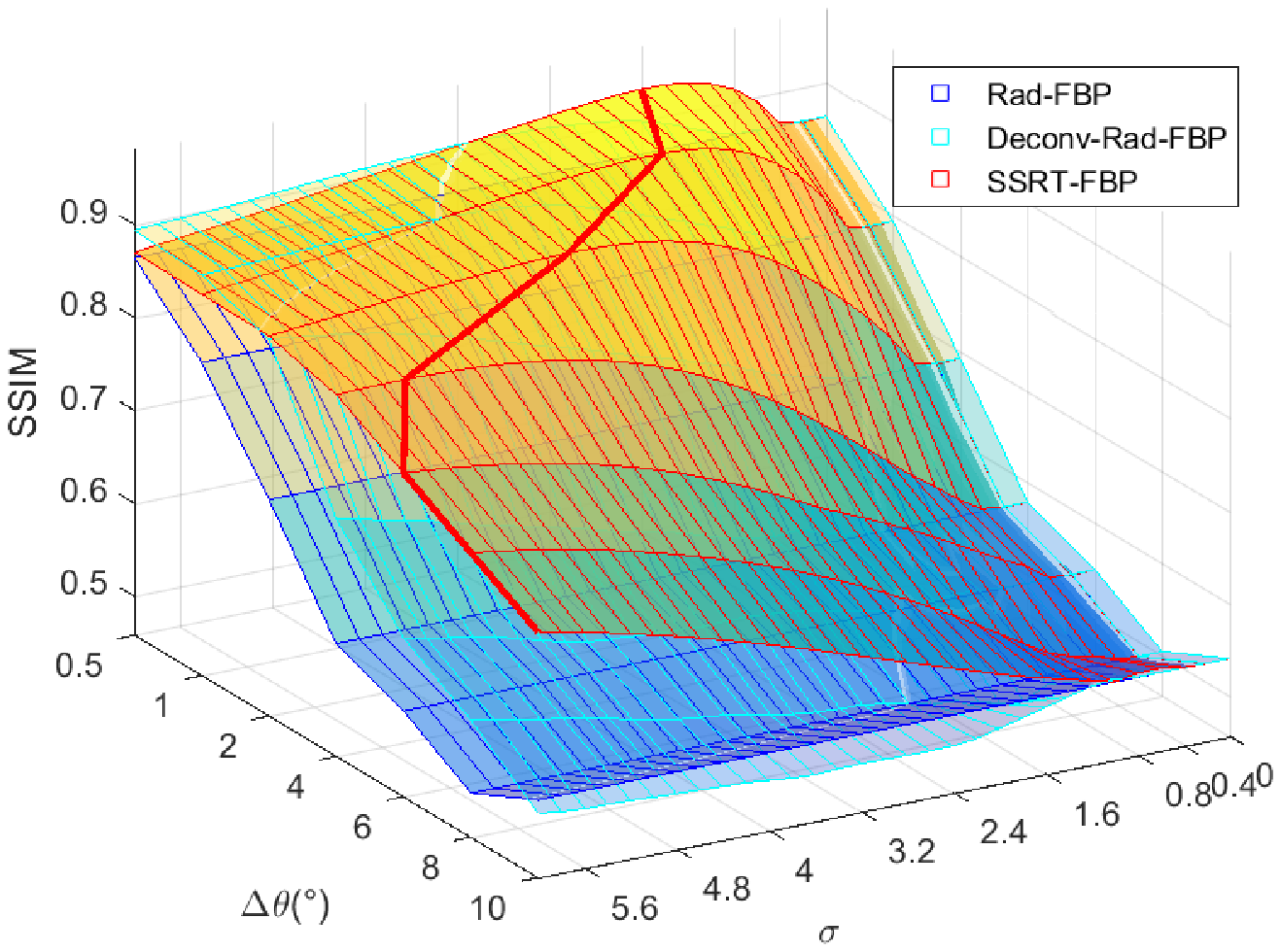}}  
    \caption{PSNR and SSIM for a sample of reconstructed abdominal phantom images in function of $\sigma$ and $\Delta\theta$. The thick red line depicts the relationship between these two parameters for which PSNR and SSIM have maximal values.} 
  \label{psnr_ssim_Liver}
\end{figure}

In Fig.~\ref{psnr_ssim_Liver}a, serving as input ground truth example, we have utilized a 512$\times$512 reconstructed image of the CT scan issued from the anthropomorphic abdominal phantom data and which depicts the 29th slice image of a total of 49 slices stacked from bottom to top, corresponding to plane A-A simulating liver arterial phase insert lesions (see Fig. 1 in \cite{Li16}).
It consists in the reconstructed anthropomorphic liver phantom with the following CT imaging protocols: (1) Acquisition parameters: (a) Scanner: 64-slice multi-detector helical CT GE 750HD , (b) peak voltage: 120 kVp, (c) effective current time product: 250 mAs, (d) pitch: 1.375; (2) Reconstruction parameters: (a) Slice thickness: 5 mm, (b) Reconstruction algorithm : FBP, (c) Convolution kernel: Standard.    
Fig.~\ref{psnr_ssim_Liver}a is used as a ground truth image against which the reconstructed images obtained from the proposed methods will be compared. It is taken as noiseless image, in terms of quantum noise, since the radiation dose of 250 mAs $\times$ 1.375 = 345 mAs (with 120 kVp tube voltage) could be considered as high, resulting in high number of incident photons per ray $I_0$, counted in billions \cite{Romero22}.         

In the following experiments, to compare the performance of the used image reconstruction methods in presence of Poisson-Gaussian noise, characterizing CT data, we generated noisy projections of anthropomorphic liver phantom with one incident exposure level $I_0=5\times 10^4$ photons per ray and one value for Gaussian noise standard deviation (std) $\sigma_n=0.5$.  

As image quality indicators, the PSNR and SSIM were calculated on the images reconstructed from the CT projections of the ground truth image shown in Fig.~\ref{psnr_ssim_Liver}a. In one case, these projections are used "as is" without any noise altering while, in an other case, the projections are corrupted by Poisson-Gaussian noise with the above-mentioned parameters. The reconstructions are done by each of the methods Radon-FBP, Deconv-Rad-FBP and SSRT-FBP and for each value of the pair ($\sigma$, $\Delta\theta$) ranging in the intervals $R_{\sigma}=[0 \ 6]$ and $R_{\Delta\theta}=[0.5\degree \ 10\degree]$, respectively. As explained in this paper, the parameters $\sigma$ and $\Delta\theta$ describe, respectively, the kernel width used for the SSRT-based methods and the number of CT projections used in image reconstruction. The results are depicted in Fig.~\ref{psnr_ssim_Liver}b-\ref{psnr_ssim_Liver}e. The thick red line depicts the optimal value of $\sigma$ for each $\Delta\theta$ for which the PSNR and SSIM, obtained for SSRT-FBP method, have maximal values, i.e. 

\[\sigma_{opt}(\Delta\theta_i)=\argmax\limits_{\sigma\in R_{\sigma}} PSNR^{SSRT-FBP}(\sigma,\Delta\theta_i), \ \forall\Delta\theta_i\in R_{\Delta\theta}\]

Similar reasoning was followed for SSIM.  

At the light of the results given in Fig.~\ref{psnr_ssim_Liver}b-\ref{psnr_ssim_Liver}e, we remark that, below a certain value of $\sigma (\sim 0.4)$, the performances of all methods are equivalent, as for the Shepp-logan phantom, where the reasons are already evoked. Above this value of $\sigma$, the image reconstruction, performed by SSRT-FBP method, is better quality, in terms of PSNR and SSIM, than the two other methods whatever the value of $\sigma$. However, for small $\Delta\theta (< 2\degree)$ (i.e. high number of CT projections), the PSNR ans SSIM values, for SSRT-FBP, are very sensitive to the variation of $\sigma$ where, maximum values could be reached by taking $\sigma$ less than 3. For SSRT-FBP method, it is clear, in this case, that the proportionality between $\sigma_{opt}$ and $\Delta\theta$ is established. This can be predictable and which can be explained as follows: when the number of the projections decreases, the lack of information can be compensated by a certain thickness of the X-ray beam projection on the CT non-zero width detector. Such a thickness is obtained by taking an appropriate $\sigma$ value of the Gaussian kernel, embedded by SSRT.  
 
When Poisson-Gaussian noise is added to the input CT data (see Figs~\ref{psnr_ssim_Liver}d and \ref{psnr_ssim_Liver}e), the similar tendency for PSNR and SSIM variations in the $\sigma-\Delta\theta$ space, as in the noiseless case (see Figs~\ref{psnr_ssim_Liver}b and \ref{psnr_ssim_Liver}c), may be seen. However, in this case, the methods Radon-FBP and Deconv-Rad-FBP have low performance, especially for PSNR, compared with SSRT-FBP, even with very small $\Delta\theta$s. This relates the sensitivity of RT to noise, in comparison to SSRT, where Gaussian kernel acts as a noise filter.   

\begin{figure}[tbh]
    \centering
    \includegraphics[scale=0.6]{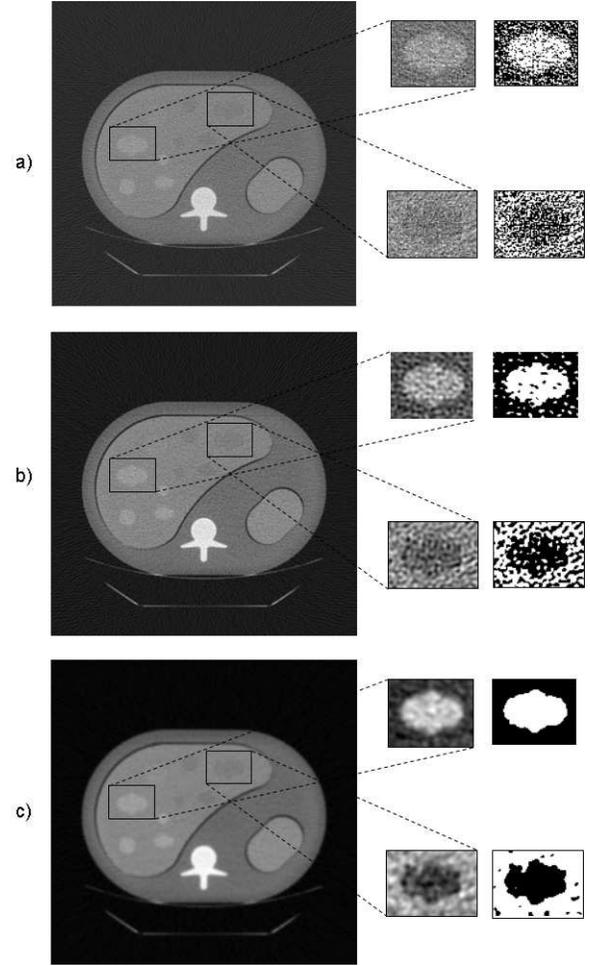}
    \caption{Image reconstruction results of the abdominal phantom (Fig.~\ref{psnr_ssim_Liver}a), with Poisson-Gaussian noise parameters $I_0=5\times 10^4$ and $\sigma_n=0.5$ for $(\sigma,\Delta\theta)=(2,1\degree)$, for the methods: (a) Radon-FBP, (b) Deconv-Rad-FBP and (c) SSRT-FBP. Zooms are done on low-contrast liver lesions with their binarizations.}
    \label{liver_recon_res} 
\end{figure}

We provide in Fig.~\ref{liver_recon_res} exhaustive reconstruction outcomes for abdominal phantom CT data with the following parameters: $(\sigma,\Delta\theta)=(2,1\degree)$ for data acquisition and $I_0=5\times 10^4$ and $\sigma_n=0.5$ for Poisson-Gaussian noise. Its ground truth image is given in Fig.~\ref{psnr_ssim_Liver}a. Figs.~\ref{liver_recon_res}a,~\ref{liver_recon_res}b and ~\ref{liver_recon_res}c depict the reconstructed images with methods of Radon-FBP, Deconv-Rad-FBP and SSRT-FBP, respectively, where the image binarization results are also given for region of interests (ROIs), inside the reconstructed images, simulating liver lesions.

It can be observed from Fig.~\ref{liver_recon_res} that visual inspection of the reconstructed images confirmed the difficulty in detecting the liver lesions or identifying the boundary of the lesions for Radon-FBP, Deconv-Rad-FBP, compared to SSRT-FBP where, here, the lesion shape is better highlighted and the noise effect is negligible. The execution time and quality measures for each of the tested methods are summarized in Table~\ref{liver_recons_resul}. 

For medical image reconstruction from the anthropomorphic abdominal phantom CT projections, these experiments ascertain again the outstanding performance of the proposed SSRT-based FBP method which meets the double challenge which are low recovery quality for low-dose CT and data noise inherent to CT systems. 

\begin{table}[tbh] 
\renewcommand{\arraystretch}{1.3}
\caption{Quality measures and runtime from image reconstruction results given in Fig.~\ref{liver_recon_res}.}
\label{liver_recons_resul} 
\centering
\begin{tabular}{c|c|c|c} 
& Radon-FBP & Deconv-Rad-FBP & SSRT-FBP \\
[2pt]\hline
PSNR & 18.02 &21.42 & \bf{27.21}\\
SSIM & 0.817 &0.867 & \bf{0.923}\\
Runtime (s) & 0.041 & 0.126 & 0.042\\
\hline
\end{tabular}
\end{table}

\section{Conclusion}
In this paper, we first presented the basic properties, the SSRT inversion approach and its application in FBP-based image reconstruction. The use of SSRT is motivated by the fact that RT, which is a line integral, does not take into account the finite dimensions of the CT system components, such as the X-ray detector. The solutions provided by some iterative ARTs such as DDM and AIM, have limited practical utility due to high computational cost. Therefore, thanks to its fast implementation, SSRT which allows accurate modeling of X-ray beam can be applied. For image reconstruction, we have originally modified the FBP algorithm to be applied on SSRT where, CT projections of Shepp-Logan and anthropomorphic abdominal phantoms, unnoisy or contaminated by Poisson-Gaussian noise, are used in experiments. The obtained results are compared with FBP-based reconstruction using two other sinograms: (1) Original RT sinogram and (2) RT sinogram derived from SSRT deconvolution. Using PSNR and SSIM as image quality metrics, the preliminary results demonstrate, in one hand, the superiority of SSRT-based image reconstruction techniques in comparison with the RT-based one and, in the other hand, the outstanding performance of SSRT-based FBP method, especially, with noisy CT data, while requiring small computation time, comparable to that of Radon-FBP. Additionally, when the number of projections is reduced, the SSRT-FBP approach reveals on  phantoms, taken as test data in experiments, to be more accurate than the two other methods, making it more appropriate for applications requiring managed radiation dose, as in medical X-ray CT.  

As future work, we plan to compare the SSRT-based CT image reconstruction with algebraic techniques using other basis functions to represent the object such as spherically symmetric volume elements or B-splines. 

\bibliographystyle{IEEEtran}
\bibliography{mybibfile}

\end{document}